\documentclass[10pt,twocolumn]{article}

\usepackage[a4paper,margin=1.9cm,columnsep=0.6cm]{geometry}

\usepackage[utf8]{inputenc}
\usepackage[T1]{fontenc}
\usepackage{lmodern}
\usepackage{microtype}

\usepackage{amsmath,amssymb,amsfonts}
\usepackage{siunitx}                 

\usepackage{graphicx}
\usepackage{booktabs}
\usepackage{caption}
\usepackage{subcaption}
\usepackage[dvipsnames]{xcolor}
\definecolor{beblue}{RGB}{0,63,114}   

\usepackage[
  colorlinks=true,
  linkcolor=beblue,
  citecolor=beblue,
  urlcolor=beblue
]{hyperref}

\usepackage{titling}
\usepackage{authblk}

\usepackage{fancyhdr}
\usepackage{abstract}

\usepackage{titlesec}
\titleformat{\section}
  {\normalfont\bfseries\large\color{beblue}}{\thesection}{0.6em}{}
\titleformat{\subsection}
  {\normalfont\bfseries\normalsize}{\thesubsection}{0.6em}{}
\titleformat{\subsubsection}
  {\normalfont\itshape\normalsize}{\thesubsubsection}{0.6em}{}
\titlespacing*{\section}{0pt}{1.4ex plus 1ex minus .2ex}{0.8ex}
\titlespacing*{\subsection}{0pt}{1.1ex plus .8ex minus .2ex}{0.6ex}

\usepackage[authoryear,round]{natbib}
\usepackage{enumitem}
\setlist{noitemsep}
\usepackage{lineno}          

\usepackage{float}
\usepackage{algorithm}
\usepackage{algpseudocode}   
\usepackage{chngcntr}        
\usepackage{placeins}

\newcommand{\keywordsname}{Keywords}
\newenvironment{keywords}{%
  \small\noindent\textbf{\keywordsname:}\ \ignorespaces
}{\par}

\title{\vspace{-1.2em}\Large\bfseries
BMCTrack-d: Pig re-identification and tracking via back marks in challenging camera settings\vspace{-0.3em}}

\small\author[1,2,*]{David Brunner}
\author[3,4]{Maciej Oczak}
\author[4]{Marie Bordes}
\author[4]{Jean-Loup Rault}
\author[1]{Stephan M. Winkler}
\author[1]{Viktoria Dorfer}

\affil[1]{Bioinformatics Research Group, PLFDoc, University of Applied Sciences Upper Austria, Softwarepark 11, Hagenberg, 4232, Austria}
\affil[2]{Computer Vision Lab, TU Wien, Favoritenstraße 9/11, Vienna, 1040, Austria}
\affil[3]{Precision Livestock Farming Hub, The University of Veterinary Medicine Vienna, Veterinärplatz 1, Vienna, 1210, Austria}
\affil[4]{Animal Welfare Science Unit, The University of Veterinary Medicine Vienna, Veterinärplatz 1, Vienna, 1210, Austria}
\affil[*]{corresponding author: \texttt{david.brunner@fh-hagenberg.at}}

\begin{document}

\date{}

\twocolumn[
  \begin{@twocolumnfalse}
    \maketitle
    \vspace{-1.5em}
    \begin{abstract}
    \noindent
Automated pig monitoring is essential for assessing their health, behaviour, and welfare. To date, most pig monitoring solutions operate on the group-level, because individual-level monitoring requires reliable long-term identification and tracking of each animal. For domesticated pigs this remains challenging because pigs of the same breed often have highly uniform appearances. Moreover, research on pig monitoring is almost exclusively reported in top-down view camera settings, which considerably ease tracking, but are not always an option in practice. In this work, BMCTrack-d is presented, a novel tracking-by-detection approach that leverages unique back marks to enable robust pig re-identification and tracking in a challenging side-view camera setting, afflicted by rapidly moving pigs, severe occlusions and low resolution. The method first predicts the detected pigs’ identities using a neural network-based back mark classifier. To improve re-identification reliability over time, two dedicated post-processing stages are introduced: a temporal prediction consistency check, which validates the identity assignments against the recent prediction history, and deduplication, which resolves conflicting identity assignments in each time step. By explicitly prioritising accurate, appearance-based re-identification over continuous tracking, the proposed approach addresses a key limitation of existing trackers for individual-level monitoring scenarios. On a demanding test set BMCTrack-d outperforms two strong baselines, BoT-SORT-ReID and TrackTrack-ReID, by 9.11\% and 1.03\%, respectively, in higher-order tracking accuracy. These results demonstrate the effectiveness of back mark-based re-identification and tracking for robust individual-level pig monitoring in challenging settings.
    \end{abstract}

    \vspace{0.6em}
    \vspace{1.2em}
  \end{@twocolumnfalse}
]

\section{Introduction}
\label{sec:intro}

Animal observation is an established way of deriving information on their reproductive state \citep{foote_estrus_1975}, health condition \citep{fernandez-carrion_motion-based_2017} and social relations \citep{clouard_evidence_2024}. Given that continuous observation imposes significant time investment on human observers and can be subjective, there is great interest in automated monitoring solutions, which typically take the form of sensors collecting data and machine learning (ML) algorithms extracting insights from these data. For the analysis of behaviour, a range of sensors can be used, like acceleration sensors \citep{mayrhuber_improved_2026} or cameras \citep{oczak_skeleton-based_2024}. The latter show better scalability, because a single camera can cover multiple animals. However, while some sensors allow differentiating animals simply by virtue of being attached to individuals, cameras necessitate advanced computer vision algorithms for this task. A number of studies have shown ML’s capacity for detecting animals \citep{liu_wheres_2023} and analysing their behaviour \citep{liu_computer_2020}. However, many of the existing studies perform behaviour recognition only on group level \citep{gan_automated_2021, gan_spatiotemporal_2022, gao_recognition_2023, li_spatiotemporal_2020, li_multi-behavior_2024, liu_computer_2020, zhang_real-time_2019}, e.g., counting the number of occurrences of certain behaviours in a pen. For specific interventions like treating a sick animal, or to identify social interactions between specific individuals (e.g., tail biting), it is necessary to monitor animals on the individual level instead.

To date, only few studies have reported results on individual-level monitoring in a group of animals \citep{odo_automated_2024, odo_re-identification_2025, t_psota_long-term_2020}. The main obstacle is the difficulty of tracking \citep{parmiggiani_dont_2023}, i.e., continuously differentiating the animals, that is required for attributing behaviours to individuals. Unlike humans, who often can be differentiated by clothing alone, animals tend to have uniform appearance, which is especially true for domesticated species like pigs, that have been intensely bred for homogenous traits to enable the standardisation of production systems. To support humans in recognising individual pigs, measures like ear marks or back marks are commonly employed, e.g., to track medical treatment or reproductive state. However, to date, only few algorithms have made use of such marks for automatic livestock monitoring \citep{fruhner_re-identifikation_2022, kashiha_automatic_2013, t_psota_long-term_2020, wutke_multistage_2025}.

Furthermore, research on pig monitoring is almost exclusively reported on top-down view camera recordings (e.g., \citet{kashiha_automatic_2013, odo_re-identification_2025, t_psota_long-term_2020}). Such a setup significantly eases tracking, e.g., by limiting the possibility of occlusions, but might be prohibitive to downstream tasks such as behaviour detection, or unfeasible due to facility constraints (see Section \ref{sec:challenges}). A side-view camera setting, in contrast, can pose significant challenges to tracking due to the possibility of severe occlusions and fluctuations in resolution whenever the pigs change their angle and distance to the camera, leading to lost identities. To enable true individual-level monitoring, a tracking algorithm must have strong re-identification (reID) capabilities to recover from such practical challenges.

Taking a step towards individual-level pig monitoring in non-idealised settings, this study proposes BMCTrack-d (back-mark-classification-based tracking with deduplication), a novel algorithm aimed at identifying and tracking pigs via their back marks, designed to prioritise strong reID over tracking in order to deal with challenging camera settings. The main contributions of this study are as follows:
\begin{itemize}
    \item BMCTrack-d, a novel algorithm for pig reID and tracking via back marks
    \item A challenging test dataset, consisting of video recordings in side-view, containing high levels of occlusion, and motion blur, as well as low-resolution pigs
    \item A comprehensive set of experiments, including of a study of BMCTrack-d’s modules and a comparison to state-of-the-art trackers
\end{itemize}

\section{Related Work}
\label{sec:related}

\subsection{Tracking}
The most widespread approach to tracking is to first detect all objects in every frame and subsequently try to match these detections across frames. This is referred to as tracking-by-detection and builds on the predictions of object detection algorithms (e.g., \citet{khanam_yolov11_2024}). In the context of deep learning, a number of high-performing methods have emerged that follow the tracking-by-detection paradigm, among which DeepSORT \citep{wojke_simple_2017}, StrongSORT \citep{du_strongsort_2023}, ByteTrack \citep{zhang_bytetrack_2022}, BoT-SORT \citep{aharon_bot-sort_2022} and most recently TrackTrack \citep{shim_focusing_2025} are prominent examples. Matching the detections across frames generally follows a two-pronged principle: 1) association by motion features and optionally 2) association by appearance features. In simple words, the algorithms try to match objects in consecutive frames by their estimated movement and optionally via their appearance. Typically, the association strategies comprise handcrafted logic like Hungarian matching, motion models and Kalman filters. Recently, \citet{gao_multiple_2025} presented MOTIP, which reformulates association as an end-to-end trainable task instead, reaching state-of-the-art results on multiple tracking benchmarks. However, contrary to previous methods which only require training data for the object detection task, end-to-end training requires a dedicated training set for the tracking task as well, making methods of this type unsuitable for scenarios with expensive data acquisition.

Interestingly, many of the studies on animal monitoring developed custom solutions instead of employing known methods. \citet{liu_computer_2020} developed a simple short-term tracking method for tracking pigs in 1-second intervals, matching pigs across frames via features derived from the detected bounding boxes. \citet{gan_automated_2021} developed an integrated architecture for the simultaneous detection and tracking of piglets. \citet{parmiggiani_dont_2023} presented an efficient tracking algorithm that matches detections across frames via a graph structure. Research that used known trackers includes \citet{lu_orp-byte_2024}, who employed a modified ByteTrack for pig tracking and \citet{guo_enhanced_2023}, who compared modified versions of several existing methods, concluding that FairMOT \citep{zhang_fairmot_2021} is best suited for the task of pig tracking, although based on outdated evaluation metrics. Tu and co-authors conducted a whole series of studies on pig tracking and behaviour detection, testing a variety of known methods: In \citet{tu_behavior_2024} and \citet{tu_tracking_2024-2} they used ByteTrack, in \citet{tu_tracking_2024} BoT-SORT and in \citet{tu_tracking_2024-1} OC-SORT \citep{maggiolino_deep_2023}.

BMCTrack-d follows the tracking-by-detection paradigm but, contrary to most other trackers, drops motion-based association in favour of appearance-based association in the shape of strong reID.

\subsection{Pig re-identification}
There are at least three ways visual pig reID has been realized in literature, via 1) ear marks, 2) back marks, and 3) marker less reID via learned appearance features. \citet{t_psota_long-term_2020} proposed an algorithm for long-term tracking of a fixed number of pigs via custom alphanumeric ear marks. \citet{fruhner_re-identifikation_2022} designed a special matrix pattern for ear marks allowing unique identification. More recently, \citet{wutke_multistage_2025} presented a multi-stage pipeline for commercial ear mark detection and identification. Given that ear marks are standard equipment in animal husbandry, using them for automated identification is a natural approach. However, many of the existing methods require custom ear marks. Furthermore, ear marks are limited in size by design, which makes them susceptible to occlusions and infeasible in case of far camera placement. Conversely, back marks can span the whole body of the pigs, making them more robust to occlusions and low resolution caused by the camera type or placement. While back marks are routinely used for manual pig identification, the only study to date to explicitly utilise back marks for automatic pig reID is \citet{kashiha_automatic_2013}, who used algorithms from classic computer vision to identify pigs via specially designed back mark patterns.

Some studies have investigated marker less pig reID \citep{odo_automated_2024, wang_towards_2022}. They use modified tracking architectures, equipped with a dedicated branch for learning to extract identifying appearance features. Some modern trackers have this capability built-in (BoT-SORT-ReID, TrackTrack-ReID). However, in both studies the pigs had distinguishing features in their appearance. The datasets of \citet{odo_automated_2024} included pigs with unique coat patterns and coloured back marks. In the study of \citet{wang_towards_2022} the pigs had numeric back marks. It stands to reason that in both cases the neural networks learned to extract these distinguishing marks, putting doubt on the feasibility of marker less pig reID. BMCTrack-d explicitly uses back marks for pig reID, outperforming methods that use learned appearance features (BoT-SORT-ReID, TrackTrack-ReID).

Moreover, all these previous studies evaluated their methods on data recorded in top-down view. The only exception is \citet{wutke_multistage_2025}, who used an additional test set consisting of side-view recordings. However, this side-view camera was set-up at close-range at a feeder and only ever shows few individuals. Conversely, the test set in this study exclusively consists of full-pen side-view recordings, allowing for severe occlusions and very low-resolution pigs ($\sim$40x30 pixels at the far end of the pen). This comprises a more realistic benchmark, given that in practice top-view cameras are not always an option, as discussed in Section \ref{sec:challenges} in more detail. Table \ref{tab:pig-reid-studies} summarizes the setup of previous studies in contrast to ours.

\begin{table*}[h]
    \centering
    \caption{Summary of existing pig reID studies.}
    \label{tab:pig-reid-studies}
    \begin{tabular}{llll}
        \toprule
        \textbf{source} & \textbf{reID modality} & \textbf{test set camera view(s)} & \textbf{resolution} \\
        \midrule
        \citet{kashiha_automatic_2013}  & back marks                          & top-down                        & 720x576              \\
        \citet{odo_re-identification_2025}      & learned appearance features         & top-down, top-down              & 1920x1080, n/a        \\
        \citet{wang_towards_2022}     & learned appearance features         & top-down                        & 1270x720              \\
        \citet{t_psota_long-term_2020}    & ear marks (custom, alphanumeric)    & top-down                        & 2688x1520             \\
        \citet{wutke_multistage_2025}    & ear marks (commercial)              & top-down, side-view (close)     & 1920x1080, n/a        \\
        \citet{fruhner_re-identifikation_2022}  & ear marks (custom, pattern)         & top-down                        & n/a                   \\
        ours                   & back marks                          & \textbf{side-view}              & 1280x720              \\
        \bottomrule
    \end{tabular}
\end{table*}

\section{Materials and Methods}
\label{sec:methods}

\subsection{Data and challenges}
\subsubsection{Experimental setup}
\label{sec:setup}
\begin{figure*}[h]
  \centering
  \includegraphics[width=1.0\textwidth]{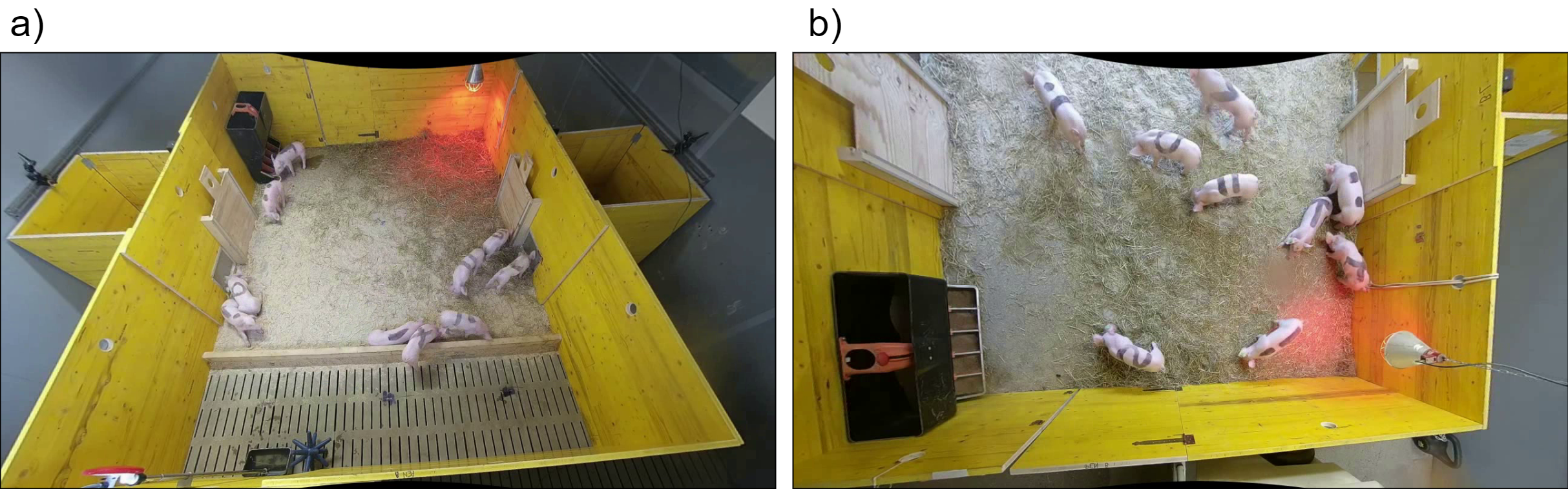}
  \caption{The camera setup. It consisted of two cameras per pen, a side-view camera (a) and a top-view camera (b). Adapted from \citet{brunner_skeleton_2026}.}
  \label{fig:dataset}
\end{figure*}

The experimental setup was identical to the one described in \citet{brunner_skeleton_2026}. It consisted of two pens (pen A, pen B), located at Medau, the pig research and teaching farm (“VetFarm”) of the University of Veterinary Medicine Vienna, Vienna, Austria.  The pens were identical in build and were constructed for an ongoing multi-national observational study on social behaviour in pigs, focusing on helping behaviour (“Let me out”, doi:10.55776/I6488). They were 3 m x 4 m in size, with a slatted area of 0.6 m x 3 m at one end and equipped with a four-head-space feeder for weaners, as well as an automatic drinker. Each pen housed exactly ten pigs (Large White x Pietrain), selected from two different litters. Daily provision of hay and food ad libitum, as well as toys for enrichment were provided. To facilitate recognition, the pigs regularly received back marks using livestock spray paint which are described in more detail in Section \ref{sec:od}. The study covered a total of seven groups of ten pigs, each group being observed for a 6-week period, from weaning (four weeks of age) to ten weeks of age. The pigs were recorded with two identical cameras (HIKVISION DS 2CD5046G0-AP, 1200x780@25, fisheye lens, Hikvision Co. Ltd., Hangzhou, Zhejiang) per pen, positioned in side view and top view, respectively. Due to the low ceiling, the top view camera covered only half of the pen. The camera streams of pen A are shown in Fig. \ref{fig:dataset}.

\subsubsection{Tracking challenges}
\label{sec:challenges}
The experimental setup described in Section \ref{sec:setup} comprises an especially challenging scenario for tracking algorithms for multiple reasons, of which the most important are 1) the camera angle, 2) the camera resolution and 3) the fast movement of the animals.

The most frequently used camera setup in pig monitoring is top view, in which the camera is placed on the ceiling facing down. This is motivated by the fact that it allows for uninterrupted observation of all pigs, mostly eliminating situations in which they cover each other. However, there are situations in which a top view camera placement is not preferrable for practical or strategic reasons. The study described in this paper provides an example for both. For one, the low ceiling of the experimental facility simply prevents obtaining useful top-view recordings. Furthermore, behavioural studies benefit from camera angles that allow observing the animals’ legs, which might be implicated in certain behaviours (e.g., lying down or pawing behaviour) but are covered in top view. A side-view camera angle, however, comes at the cost of the aforementioned occlusions, caused by pigs covering their mates from the perspective of the camera. Occlusions can lead to lost tracks and pose an important challenge for tracking.

The moderate resolution of the cameras used in this study (1200 x 780 pixels) is another challenge for tracking algorithms. While high resolution cameras have become relatively inexpensive, their recordings take up much more disk space, which, in the case of continuous recordings, can be prohibitive. Also, because automated monitoring of animals is still a nascent field and previous studies were conducted primarily with human observers in mind, which are less encumbered by low resolution than algorithms, it should be expected that much available video material is in low resolution. In low resolution, physical characteristics that would allow differentiating individual animals are less pronounced, posing a challenge for trackers that use appearance features in addition to movement.

\begin{figure*}[h]
  \centering
  \includegraphics[width=0.85\textwidth]{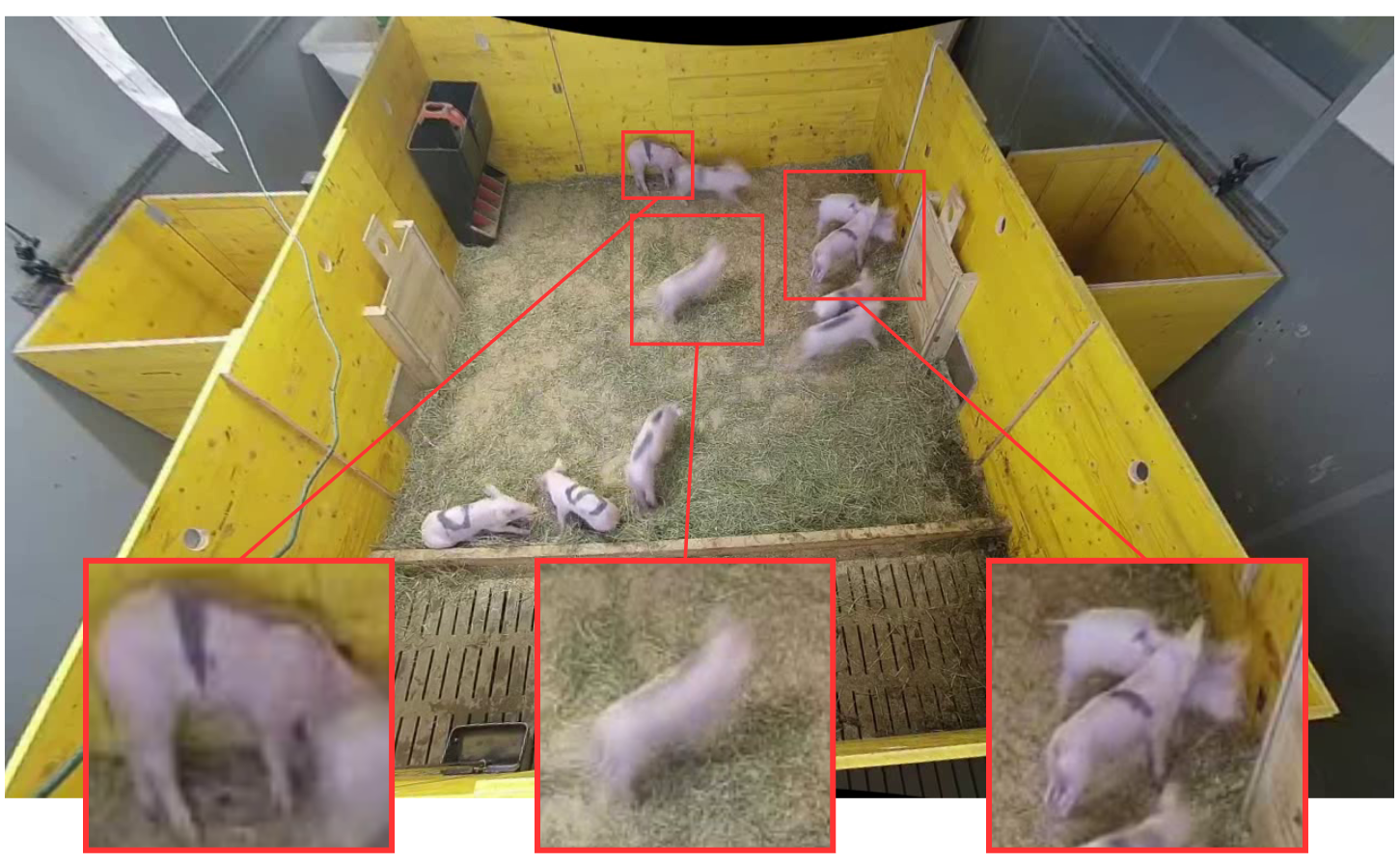}
  \caption{The challenges posed by the experimental setup of the study. From left to right: low resolution, motion blur and occlusions.}
  \label{fig:challenges}
\end{figure*}

Lastly, the study described in this paper is focused on young pigs, which are very active and prone to fast movement, which can pose a challenge for tracking algorithms for two reasons. First, large jumps in location make it hard to associate an individual in consecutive frames via its position. Second, tracking algorithms that rely on physical appearance are also affected by motion blur, which, similarly to low resolution, weakens idiosyncratic features that could be used for differentiation. Figure \ref{fig:challenges} illustrates the described challenges.

\subsection{Proposed tracking algorithm}
\subsubsection{Overview}
\begin{figure*}[h]
  \centering
  \includegraphics[width=0.85\textwidth]{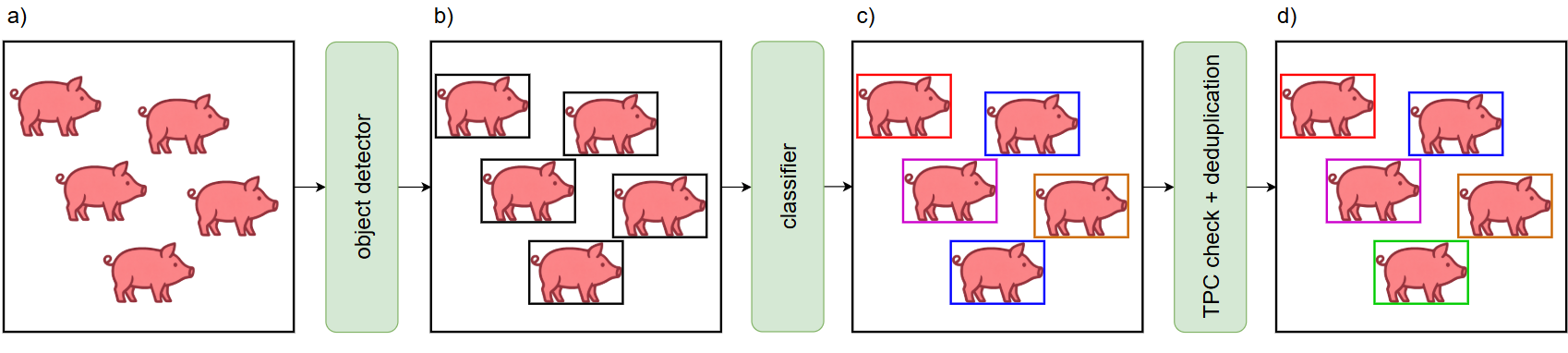}
  \caption{Overview of the workflow of BMCTrack-d. For a given input frame (a), object detection is performed to localize the pigs (b), subsequent classification predicts their identity (c) and a prediction refinement step consisting of a temporal prediction consistency check and deduplication is performed to improve the identification (d).}
  \label{fig:overview}
\end{figure*}

The proposed tracking algorithm is termed BMCTrack-d (back-mark-classification-based tracking with deduplication) and consists of four steps: 1) object detection, 2) back mark classification, 3) temporal prediction consistency (TPC) check and 4) deduplication. For each frame, first, bounding boxes around all pigs in the scene are inferred. Then, each bounding box area is passed to an image classifier that predicts which individual is shown via its back mark. Next, the class predictions are refined by checking if they are consistent with the predictions in previous frames. Finally, by ensuring that each class is uniquely represented in the frame, remaining class collisions are resolved. Figure \ref{fig:overview} shows a high-level illustration of the whole algorithm. Sections \ref{sec:od} – \ref{sec:matching} provide details on the individual components.

\subsubsection{Object detection and back mark classification}
\label{sec:od}
BMCTrack-d follows the tracking-by-detection paradigm, using YOLOv11 \citep{khanam_yolov11_2024} as the object detector. While in this first detection step the algorithm closely follows existing trackers (e.g., ByteTrack \citep{zhang_bytetrack_2022}, BoT-SORT \citep{aharon_bot-sort_2022}), the next step is where it diverges. Tracking algorithms typically assign a random ID to each detected object and then try to preserve this association between object and ID throughout the video clip \citep{aharon_bot-sort_2022, du_strongsort_2023, maggiolino_deep_2023, wojke_simple_2017, zhang_bytetrack_2022}. Given that for the described behavioural study, it is important not only to keep the individuals apart over time but also to know their identity, a separate classification step is added. The detected bounding boxes are cropped from the frame and passed to an image classifier of type ResNet-50 \citep{he_deep_2016}, which predicts which of a list of known individuals the detected pig represents. Performed in every frame, this results in a track for each individual. 

\begin{figure*}[h]
  \centering
  \includegraphics[width=0.85\textwidth]{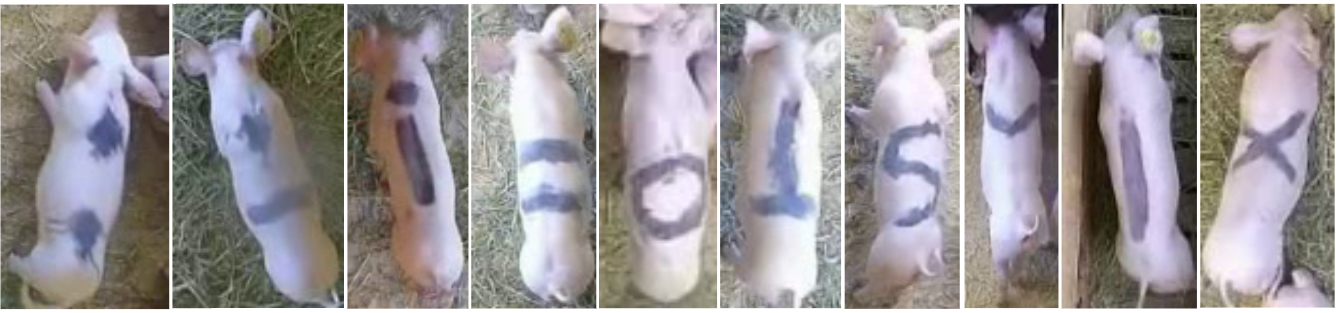}
  \caption{An example for each back mark used in this study. From left to right: \textit{dot dot}, \textit{dot line horizontal}, \textit{i}, \textit{line line horizontal}, \textit{o}, \textit{reverse t}, \textit{s}, \textit{v}, \textit{vertical line}, \textit{x}. Adapted from \citet{brunner_insights_2026}.}
  \label{fig:backmarks}
\end{figure*}

To support recognition, the pigs were equipped with back marks. Figure \ref{fig:backmarks} shows the ten unique back marks used in this study. They were renewed every 3-4 days to ensure good readability. As the back marks were hand drawn, they were not perfectly consistent across groups. Insights into ways of improving the back mark design for future studies are described in \citep{brunner_insights_2026}.

\subsubsection{Temporal prediction consistency}
\label{sec:tpc}
\begin{figure*}[h]
  \centering
  \includegraphics[width=0.85\textwidth]{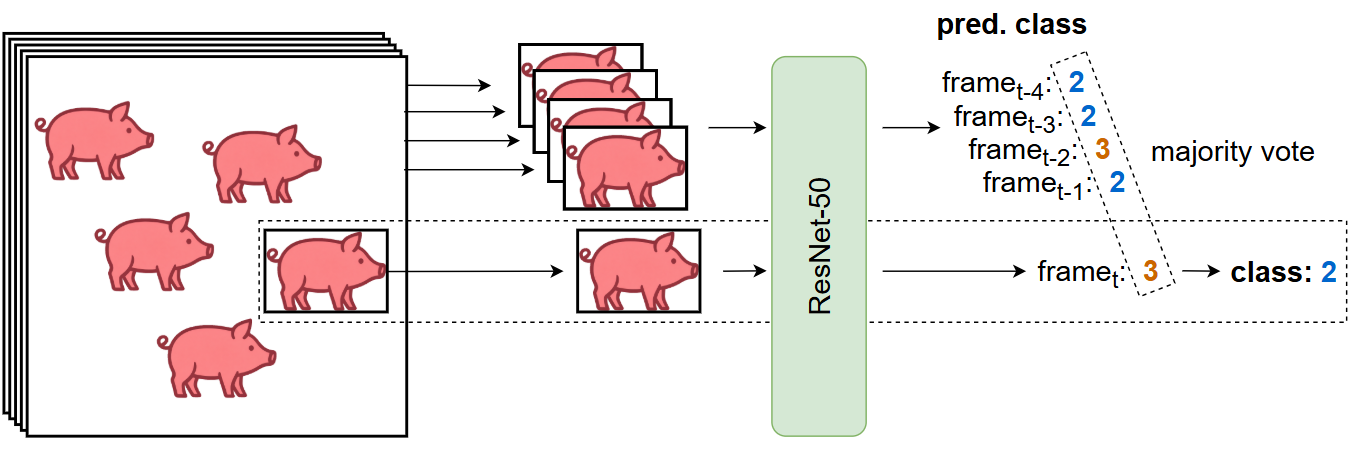}
  \caption{An illustration of the temporal prediction consistency check. The current prediction is not in line with the previous $n$ predictions, so it is updated to the majority prediction.}
  \label{fig:tpc}
\end{figure*}

Recognising the individuals in any given frame requires the back marks to be visible. As described in Section \ref{sec:challenges}, the data produced in this study include frames with occlusions, motion blur and moderate resolution, all of which pose obstacles for recognition. Due to the side-view camera angle, the visibility of the back marks is also dependent on the body pose and the orientation of the pigs in relation to the camera. Therefore, it is to be expected that the recognition in individual frames contains errors. To refine the class predictions for a given frame, they are compared to the associated predictions in previous frames. Thus, it is not necessary for each class prediction to be correct in a sequence of frames, only that the majority of predictions are correct. By ensuring the temporal prediction consistency (TPC) of a sequence of predictions, the tracking accuracy can be improved. Figure \ref{fig:tpc} illustrates this idea. Specifically, the class ID of the current frame is determined by the majority class prediction in the last $n$ frames (the \textit{frame buffer}) and the current prediction. If there is no majority, the current prediction is kept. The TPC check starts once $n$ frames have accumulated. Section \ref{sec:example} gives a concrete example for this process.

\subsubsection{Deduplication}
\begin{figure*}[h]
  \centering
  \includegraphics[width=0.85\textwidth]{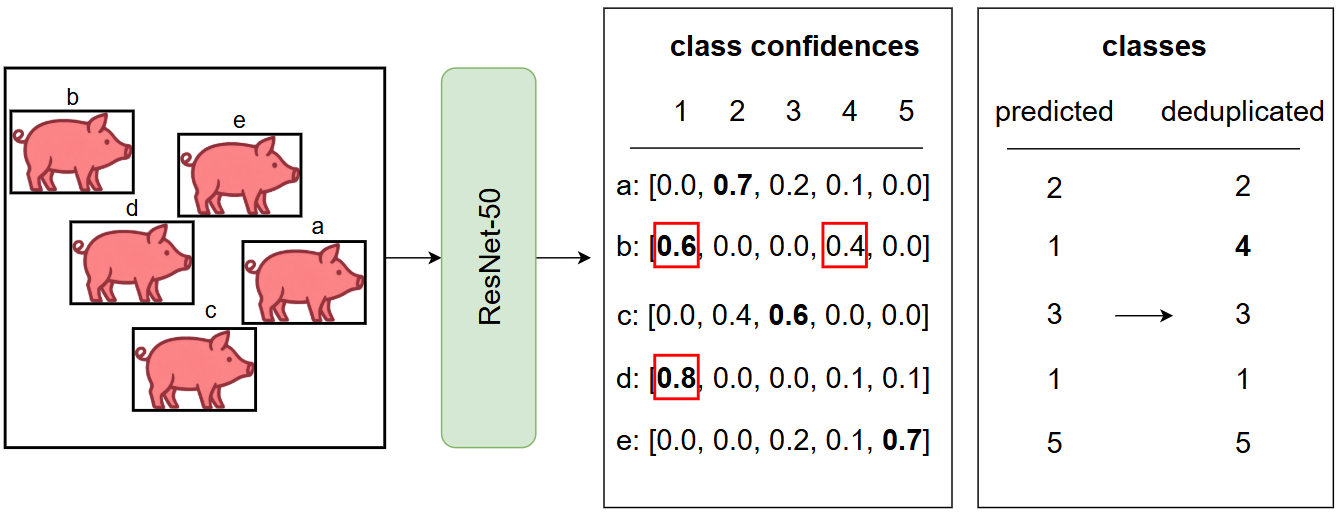}
  \caption{An illustration of the deduplication. The ResNet-50 classifier produces a confidence vector for each pig, indicating which class it most likely belongs to. Pigs $b$ and $d$ are both assigned to class 1. However, $b$ has a lower maximum confidence for class 1 and can be reassigned to its second most likely class 4. For simplicity a scenario with a total of five pigs is assumed.}
  \label{fig:deduplication}
\end{figure*}

The classifications of individual bounding boxes in a frame are independent of each other, meaning that the assigned class IDs are not unique and multiple pigs can be assigned identical class IDs. Given that each individual can only appear once per frame, these collisions in recognition must be resolved. The object detection step is ignorant of individual class IDs but provides confidence scores for each predicted bounding box. As the number of pigs per pen is known to be exactly ten, in a first step, the bounding boxes can be ranked by confidence and clipped to ten. Unlike the object detector, the classifier produces a vector including a confidence score for each class ID. If collisions occur among the remaining bounding boxes, i.e., if two predictions assign the highest confidence to the same class, the one with the lower maximum confidence can be shifted to its second highest confidence class as illustrated in Fig. \ref{fig:deduplication}. Given that this could result in a new collision, the process is repeated until each class prediction is unique. This deduplication is performed after the TPC check and overrides the latter.

\subsubsection{Matching variants}
\label{sec:matching}
\begin{figure*}[h]
  \centering
  \includegraphics[width=0.85\textwidth]{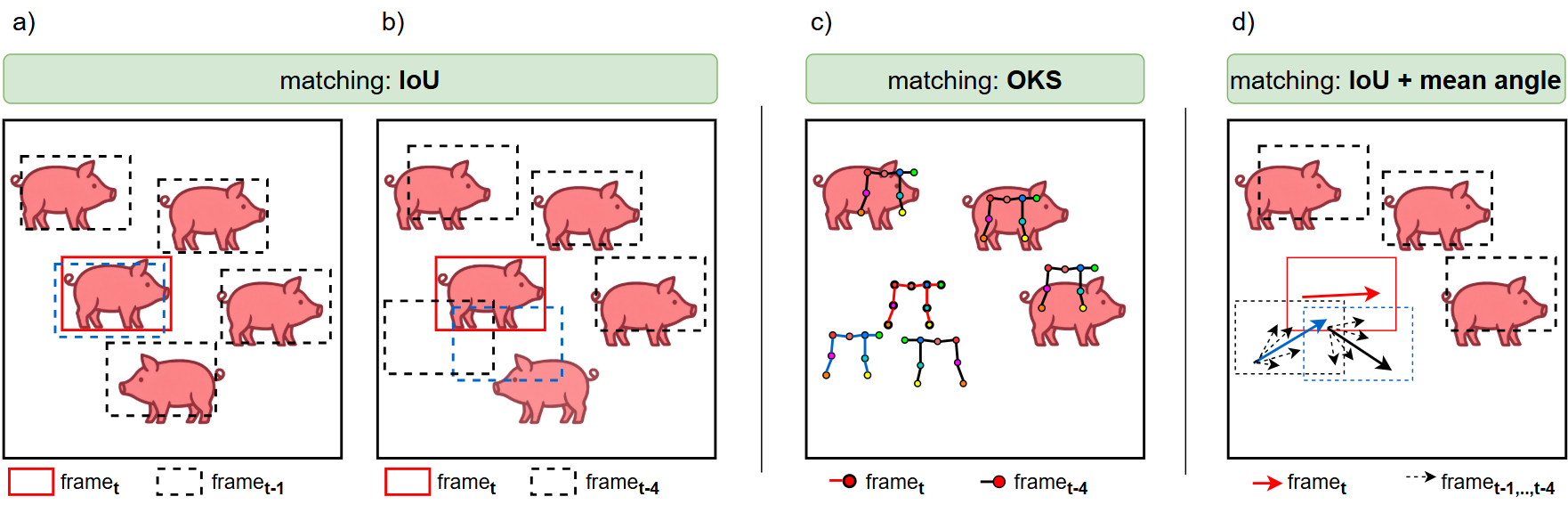}
  \caption{The matching variants. Overlap-based matching works well for consecutive frames (a) but might produce wrong results across larger gaps (b). Alternatively, keypoint skeletons (c) or the pigs’ orientation (d) can be used for matching instead. The additional information of pose and orientation helps to recover the correct match. Blue boxes, skeletons and arrows indicate a match.}
  \label{fig:matching}
\end{figure*}

The TPC check described in Section \ref{sec:tpc} compares class predictions over time. In order to compare the class prediction for a specific individual in frame $t$ with its predecessors in the last $n$ frames, some sort of matching has to be performed to find the same individual in previous frames. One way of doing so is to measure the Intersection over Union (IoU) of all bounding boxes in two sequential frames and assume that those with the highest IoU represent the same individual. This typically works well for consecutive frames, as shown in Fig. \ref{fig:matching}a. However, depending on the video frame rate and the speed of the animals’ movement, it can fail for bigger temporal jumps, e.g., matching a bounding box in frame $t$ with those in frame $t$-4, as illustrated in Fig. \ref{fig:matching}b. An alternative to matching bounding boxes is matching the keypoint skeletons predicted by a pose estimation model, as illustrated in Fig. \ref{fig:matching}c. The advantage of matching keypoint skeletons is that they not only encode the pigs’ locations but also their pose and orientation. If a pig in frame $t$ occupies the same location that a different pig occupied in frame $t$-4, but they differ in their pose or orientation, a wrong match can be avoided. Analogously to IoU for matching bounding boxes, object keypoint similarity\footnote{https://cocodataset.org/\#keypoints-eval} (OKS) serves as metric for matching keypoint skeletons. Figure \ref{fig:matching}d depicts a simpler alternative to full skeleton matching, in which IoU-based matching is supplemented with a comparison of the current orientation of the pig in frame $t$ to the mean orientation of the matched pig instances in the previous $n$ frames. The orientation can be derived from only two keypoints, one at each end of the pig’s body. ViTPose \citep{xu_vitpose_2022} serves as the pose estimation model; for details on model training, data and the keypoint skeleton structure refer to \citet{brunner_skeleton_2026}.

\subsubsection{Algorithmic details}
\label{sec:example}
\begin{figure*}[h]
  \centering
  \includegraphics[width=0.85\textwidth]{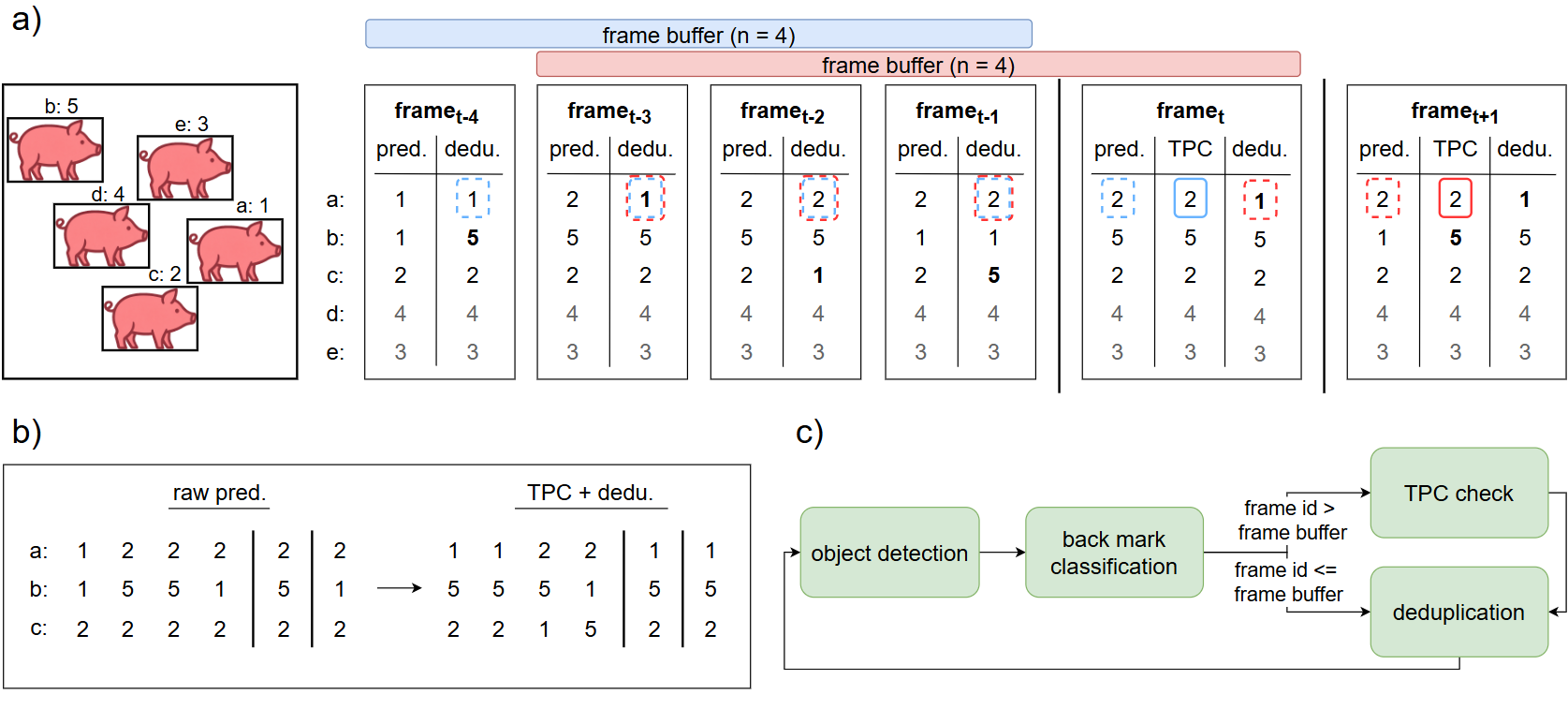}
  \caption{Details on BMCTrack-d’s algorithm. The processing of the first six frames of a video clip (a) is exemplified, as well as the resulting adjusted tracks (b) and the general workflow of the algorithm (c). For simplicity a scenario with a total of five pigs is assumed.}
  \label{fig:example}
\end{figure*}

For a better understanding of how BMCTrack-d works, this section provides a concrete example, showcasing the interplay between the steps. Figure \ref{fig:example}c illustrates the general workflow. The TPC check starts only after the number of frames specified by the frame buffer length were processed. Up until this point, for each frame only object detection, followed by back mark classification and deduplication are performed. Once the frame buffer is full, the TPC check is added between the back mark classification and deduplication. Figure \ref{fig:example}a sketches the processing of the first six frames of a video clip showing five pigs with unique back marks, for frame buffer length $n=4$. In frame $t$-4, after object detection and back mark classification, duplicate predictions for pigs $a$ and $b$ are resolved by deduplication, switching $b$’s class ID to 5. The same principle applies to frames $t$-3 to $t$-1, the changed predictions in bold. From frame $t$ forward, the TPC check is added. The sequence of predictions in previous frames (blue dashed boxes) reads 1, 1, 2, 2, the prediction in frame $t$ (blue solid box) is 2. The latter makes class 2 the majority prediction and it is adopted for frame $t$. However, because the prediction confidence (not shown in the figure) of class 2 for pig $a$ is lower than for pig $c$, deduplication switches pig $a$’s class to 1, overriding the TPC check. In frame $t$+1 pig $a$ is, again, misclassified as 2, which results in a majority in the sequence (red boxes) and the adoption of 2 by the TPC check, an error, once more corrected by deduplication. Figure \ref{fig:example}b compares the raw class predictions to the final corrected predictions by BMCTrack-d and shows a trend in which the tracks of pigs $a$ and $b$ are gradually corrected to their true classes 1 and 5 at the cost of some errors in pig $c$’s track. Pigs $d$ and $e$ are detected correctly and also unaffected by deduplication in this example.

\subsection{Experiments}
\label{sec:experiments}

\subsubsection{Datasets and model training}
\label{sec:datasets}
The training and validation datasets for the object detector consist of 567 frames and 30 frames, respectively, which were extracted from the video recordings collected in the study. They span multiple groups of pigs, in both pens (pen A, pen B) and both camera views (side view, top view). A summary is provided in Table \ref{tab:yolo-datasets}. All 5715 pig instances were annotated with bounding boxes using either the Computer Vision Annotation Tool\footnote{https://app.cvat.ai}  (CVAT) or COCO Annotator \citep{brooks_coco_2019}. The YOLOv11 object detector reached 99.42\% mean average precision at IoU threshold 0.5 (mAP@0.5) on the validation set. For the training of the classifier a different, but overlapping set of the data was used, the areas of the annotated bounding boxes were extracted and all crops in which the back marks were not visible manually filtered. This version of the training and validation datasets, as well as the classifier training are explained in more detail in \citet{brunner_insights_2026}. The classifier reached 91\% accuracy on the validation set. Both the object detector’s and the classifier’s validation data were sampled from held out video data, recorded independently from the training data. This is to ensure generalisation across visual properties that differ over time, such as variations in the back marks, which cannot be drawn identically every time.  The dataset and training details of the ViTPose pose estimation model are described in \citet{brunner_skeleton_2026}.\\
\begin{table*}[h]
    \centering
    \caption{The number of frames in the training and validation datasets for the YOLOv11 object detector.}
    \label{tab:yolo-datasets}
    \begin{tabular}{lcccccc}
        \toprule
        \textbf{dataset} & \textbf{total} & \textbf{side view} & \textbf{top view} & \textbf{pen A} & \textbf{pen B} & \textbf{instances total} \\
        \midrule
        training   & 567 & 454 & 113 & 396 & 171 & 5443 \\
        validation & 30  & 24  & 6   & 15  & 15  & 272  \\
        \bottomrule
    \end{tabular}
\end{table*}
The test dataset for evaluating BMCTrack-d consists of a total of 10 video clips, between 10 and 30 seconds in duration. The clips span multiple groups of pigs, show both pens and amount to 3500 frames in total. They exclusively consist of recordings from the more challenging side view angle and were annotated with bounding boxes and associated class IDs using CVAT. The clips were selected, such that they cover a range of scenarios which are relevant for tracking, as detailed in Table \ref{tab:clip-overview}. This range of scenarios could not have been achieved by sampling from a single group, hence there was no dedicated, held-back test group. The loose categorisation into the three difficulty classes (low, medium, hard) in Table \ref{tab:clip-overview} indicates the expected difficulty of the test clips, dictated by properties such as speed of movement, average distance to camera and occlusions throughout the clips. The test clips are short because pigs tend to exhibit long stretches of stationary behaviour, interrupted by moments of interaction and bouts of frantic movement. It is precisely these latter situations that are most demanding on a tracker’s ability to retain tracks and hence provide the most information about a tracker’s performance. A visualisation of the pigs’ movement trajectories per clip can be found in Appendix \ref{app:vis}, Fig. \ref{app:datatracks}.\\
\begin{table*}[h]
    \centering
    \caption{Overview of video clips used for evaluation.} 
    \label{tab:clip-overview}
    \begin{tabular}{llllp{7cm}}
        \toprule
        \textbf{clip} & \textbf{\# frames} & \textbf{duration (s)} & \textbf{difficulty} & \textbf{properties} \\
        \midrule
        penA\_10s\_5 & 250 & 10 & medium & low movement; dispersed; moderate back mark visibility \\
        penA\_10s\_6 & 250 & 10 & low    & moderate movement; dispersed; good back mark visibility \\
        penA\_10s\_7 & 250 & 10 & medium & moderate movement; dispersed; moderate back mark visibility \\
        penA\_10s\_8 & 250 & 10 & medium & moderate movement; increased distance to camera; moderate back mark visibility \\
        penA\_30s\_2 & 750 & 30 & high   & very fast movement; dispersed; severely blurred and occluded back marks \\
        penA\_30s\_5 & 750 & 30 & high   & very fast movement; dispersed; severely blurred and occluded back marks \\
        penB\_10s\_2 & 250 & 10 & low    & moderate movement; close to camera; good back mark visibility \\
        penB\_10s\_5 & 250 & 10 & low    & low movement; dispersed; good back mark visibility \\
        penB\_10s\_6 & 250 & 10 & medium & low movement; increased distance to camera; moderate back mark visibility \\
        penB\_10s\_7 & 250 & 10 & medium & moderate movement; increased distance to camera; moderate back mark visibility \\
        \bottomrule
    \end{tabular}
\end{table*}
All experiments were run on Ubuntu 20.04 and a NVIDIA GeForce RTX 3090 GPU (NVIDIA driver version 535.171.04., CUDA version: 12.2). The code was developed in Python 3.10 and the PyTorch framework.

\subsubsection{Tracking metrics}
The metric adopted for the evaluation of the tracking algorithms in this work is the higher order tracking accuracy (HOTA) \citep{luiten_hota_2021}. HOTA can be decomposed into the sub-metrics localisation accuracy (LocA), association accuracy (AssA), and detection accuracy (DetA) for a more nuanced analysis of the tracking performance. A detailed description of the HOTA framework can be found in Appendix \ref{app:metrics}.

\subsubsection{Optimal configuration study}
In the first set of experiments several variants of the proposed tracking algorithm are evaluated. To quantify the benefits of BMCTrack-d’s individual components, an ablation study is performed, discarding the TPC-check (BMC-d), deduplication (BMCTrack) or both (BMC). For all variants the frame buffer length is set $n=4$ and IoU used for matching. Additional experiments on the optimal frame buffer length, alternative matching variants and oriented bounding boxes can be found in Appendix \ref{app:additional}.

\subsubsection{Algorithm evaluation}
The second set of experiments pits BMCTrack-d against two established trackers and one novel tracker. ByteTrack \citep{zhang_bytetrack_2022} is an efficient tracker, which achieves high tracking performance by introducing the idea of re-matching low-confidence detections in a second matching round. BoT-SORT \citep{aharon_bot-sort_2022} refines the ByteTrack algorithm by a more accurate Kalman filter forecasting, as well as camera motion compensation. TrackTrack \citep{shim_focusing_2025} introduces a track-centric matching strategy that assigns detections to existing tracks, rather than globally associate detections into tracks. Contrary to ByteTrack, which solely operates on location-based detection matching, BoT-SORT and TrackTrack offer optional appearance-based ReID capabilities. To this end, an appearance feature vector is extracted from the bounding box of a detection with a deep neural network. A detection in a subsequent frame is matched, if the cosine similarity of the appearance feature vector with the new detection’s appearance features falls below a set threshold. The appearance feature vector is then updated by the features of the matched detection via the exponential moving average mechanism. This technique allows reinstating lost tracks, by using appearance clues to recognise that a recently disappeared and newly appearing individual are in fact the same. However, because of the continuous updating of the appearance vector this capability is typically limited to short time windows (e.g., 30 frames).
The object detector is identical for all trackers. To facilitate a fair comparison, the strategy of clipping the set of predicted bounding boxes to ten per frame, employed in BMCTrack-d, is adopted for the other trackers as well. As these trackers are oblivious to the back mark classes and assign numeric IDs instead, for evaluation purposes it is assumed that their predictions in the first frame are perfect and a static mapping between the back mark classes and the assigned IDs was created. Finally, a runtime efficiency evaluation is performed, comparing the latency and throughput of the methods.

\section{Results}
\label{sec:results}

\subsection{Optimal configuration study}
The results of the ablation study on algorithm variants are presented in Table \ref{tab:ablation-study}. They show the baseline (BMCTrack-d) to outperform all variants (BMC, BMC-d, BMCTrack). Section \ref{sec:discussion} discusses the results in more detail.

\begin{table}[h]
    \centering
    \caption{The results of the ablation study. Best results in bold.}
    \label{tab:ablation-study}
    \begin{tabular}{lcccc}
        \toprule
        \textbf{method} & \textbf{LocA} & \textbf{AssA} & \textbf{DetA} & \textbf{HOTA} \\
        \midrule
        BMC              & 0.8864 & 0.4374          & 0.7388 & 0.5633          \\
        BMC-d            & 0.8864 & 0.5667          & 0.8591 & 0.6901          \\
        BMCTrack   & 0.8864 & 0.4056          & 0.7232 & 0.5261          \\
        BMCTrack-d & 0.8864 & \textbf{0.7048} & 0.8591 & \textbf{0.7714} \\
        \bottomrule
    \end{tabular}
\end{table}

\subsection{Algorithm evaluation}
Table \ref{tab:tracking-evaluation} presents a comparison of the test set performances between BMCTrack-d, ByteTrack, BoT-SORT and TrackTrack. It shows BMCTrack-d to outperform the other trackers, often by a significant margin. While ByteTrack remains non-competitive on all scores, BoT-SORT performs nearly identically to BMCTrack-d w.r.t. the localisation (LocA) and detection (DetA) but stays behind on the association score (AssA). TrackTrack further closes the gap, especially on enabling the ReID capabilities. Tables \ref{tab:per-clip-bmctrack-d} and \ref{tab:per-clip-tracktrack-reid} show the performances on each of the test set clips separately for BMCTrack-d and TrackTrack, respectively. Both trackers’ performances show a similar trend on most clips but diverge significantly on \textit{penA\_30s\_2}. The results are examined in detail in Section \ref{sec:discussion}. Table \ref{tab:runtime-efficiency} shows the results of the runtime efficiency evaluation.

\begin{table}[h]
    \centering
    \caption{The results of the tracking algorithm evaluation. Best results in bold.}
    \label{tab:tracking-evaluation}
    \resizebox{\columnwidth}{!}{%
        \begin{tabular}{lcccc}
            \toprule
            \textbf{method} & \textbf{LocA} & \textbf{AssA} & \textbf{DetA} & \textbf{HOTA} \\
            \midrule
            ByteTrack        & 0.8531          & 0.6069          & 0.7915          & 0.6680          \\
            BoT-SORT         & 0.8843          & 0.6525          & 0.8581          & 0.7173          \\
            BoT-SORT-ReID    & 0.8841          & 0.5847          & 0.8576          & 0.6803          \\
            TrackTrack       & \textbf{0.8877} & 0.6692          & 0.8583          & 0.7356          \\
            TrackTrack-ReID  & \textbf{0.8877} & 0.6970          & 0.8586          & 0.7611          \\
            BMCTrack-d       & 0.8864          & \textbf{0.7048} & \textbf{0.8591} & \textbf{0.7714} \\
            \bottomrule
        \end{tabular}
    }
\end{table}

\begin{table}[h]
    \centering
    \caption{Per-clip results of BMCTrack-d on the test set.}
    \label{tab:per-clip-bmctrack-d}
    \begin{tabular}{lcccc}
        \toprule
        \textbf{clip} & \textbf{LocA} & \textbf{AssA} & \textbf{DetA} & \textbf{HOTA} \\
        \midrule
        penA\_10s\_5 & 0.7694 & 0.4013 & 0.7635 & 0.5535 \\
        penA\_10s\_6 & 0.9152 & 0.8855 & 0.8855 & 0.8855 \\
        penA\_10s\_7 & 0.8957 & 0.8503 & 0.8611 & 0.8557 \\
        penA\_10s\_8 & 0.9064 & 0.4917 & 0.8938 & 0.6624 \\
        penA\_30s\_2 & 0.8676 & 0.5162 & 0.8156 & 0.6486 \\
        penA\_30s\_5 & 0.8808 & 0.4686 & 0.8428 & 0.6282 \\
        penB\_10s\_2 & 0.9003 & 0.8832 & 0.8854 & 0.8843 \\
        penB\_10s\_5 & 0.9149 & 0.8969 & 0.8969 & 0.8969 \\
        penB\_10s\_6 & 0.9020 & 0.8699 & 0.8819 & 0.8759 \\
        penB\_10s\_7 & 0.9122 & 0.7841 & 0.8645 & 0.8233 \\
        \midrule
        mean         & 0.8864 & 0.7048 & 0.8591 & 0.7714 \\
        \bottomrule
    \end{tabular}
\end{table}

\begin{table}[h]
    \centering
    \caption{Per-clip results of TrackTrack-ReID on the test set.}
    \label{tab:per-clip-tracktrack-reid}
    \begin{tabular}{lcccc}
        \toprule
        \textbf{clip} & \textbf{LocA} & \textbf{AssA} & \textbf{DetA} & \textbf{HOTA} \\
        \midrule
        penA\_10s\_5 & 0.7734 & 0.6416 & 0.7603 & 0.6981 \\
        penA\_10s\_6 & 0.9157 & 0.8325 & 0.8853 & 0.8584 \\
        penA\_10s\_7 & 0.8973 & 0.8356 & 0.8620 & 0.8487 \\
        penA\_10s\_8 & 0.9066 & 0.8944 & 0.8944 & 0.8944 \\
        penA\_30s\_2 & 0.8699 & 0.1688 & 0.8136 & 0.3704 \\
        penA\_30s\_5 & 0.8824 & 0.4412 & 0.8415 & 0.6093 \\
        penB\_10s\_2 & 0.9007 & 0.8631 & 0.8872 & 0.8750 \\
        penB\_10s\_5 & 0.9150 & 0.8981 & 0.8981 & 0.8981 \\
        penB\_10s\_6 & 0.9025 & 0.7282 & 0.8825 & 0.8015 \\
        penB\_10s\_7 & 0.9131 & 0.6661 & 0.8614 & 0.7574 \\
        \midrule
        mean         & 0.8876 & 0.6970 & 0.8586 & 0.7611 \\
        \bottomrule
    \end{tabular}
\end{table}

\begin{table}[h]
    \centering
    \caption{The results of the runtime efficiency evaluation. All experiments were run on a NVIDIA GeForce RTX 3090 GPU.}
    \label{tab:runtime-efficiency}
    \resizebox{\columnwidth}{!}{%
    \begin{tabular}{lcc}
        \toprule
        \textbf{method} & \textbf{mean latency (ms)} & \textbf{fps} \\
        \midrule
        ByteTrack                         & \textbf{16.79}          & \textbf{59} \\
        BoT-SORT-ReID                     & 31.11          & 32 \\
        TrackTrack-ReID                   & 31.88          & 31 \\
        BMCTrack-d (of which TPC + dedup.) & 143.48 (1.12) & 7  \\
        \bottomrule
    \end{tabular}}
\end{table}

\section{Discussion}
\label{sec:discussion}

\subsection{Strengths and weaknesses of BMCTrack-d}
\begin{figure*}[h]
  \centering
  \includegraphics[width=1.0\textwidth]{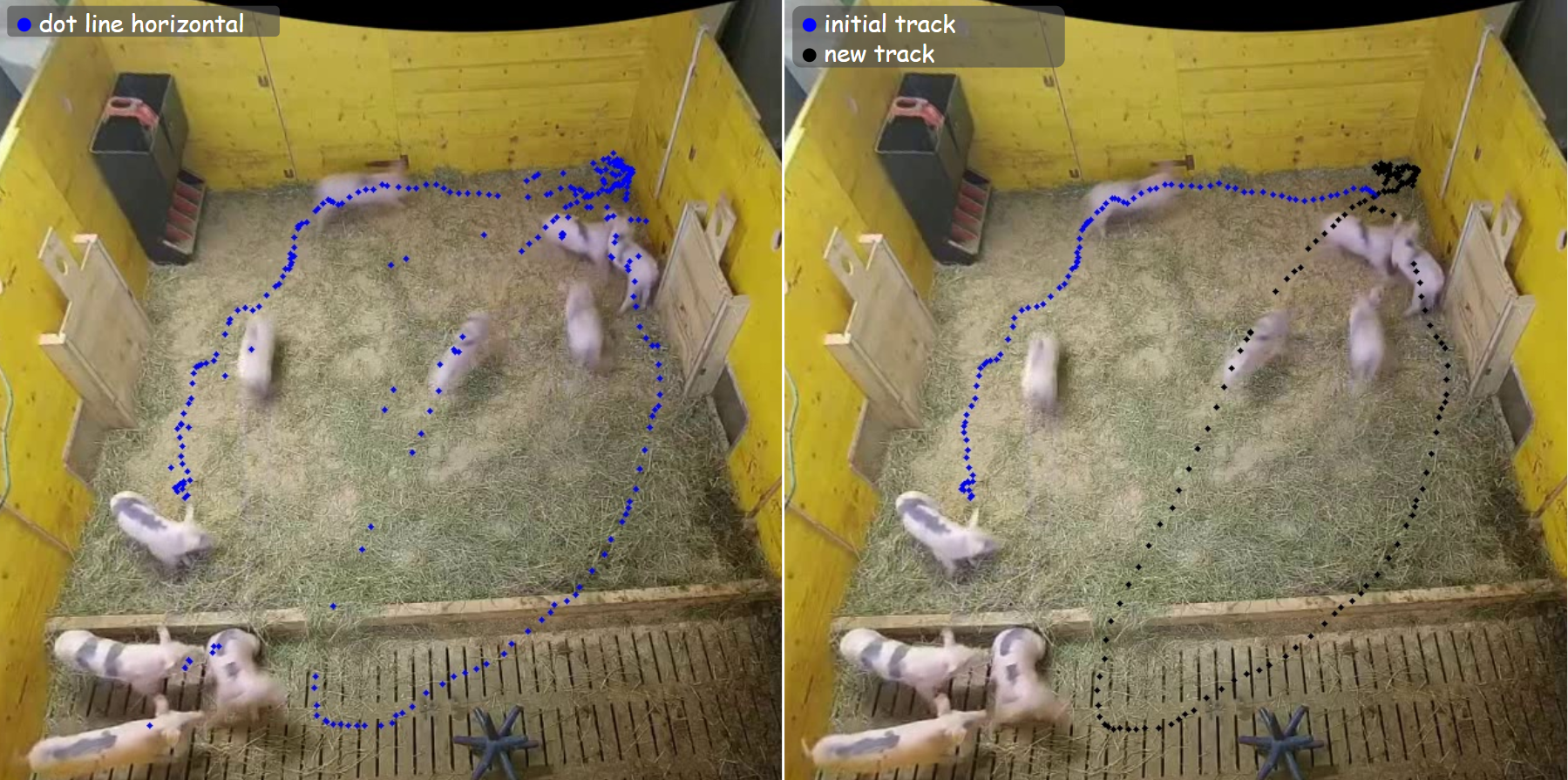}
  \caption{Qualitative comparison between BMCTrack-d (left) and BoT-SORT (right) for tracking a single pig. The blue dots show the detected locations of \textit{dot line horizontal} throughout the video. The black dots in the right image signalize the track ID changing, i.e., BoT-SORT losing the track.}
  \label{fig:qualitative}
\end{figure*}

While some existing trackers can be described as tracking with ReID capabilities \citep{aharon_bot-sort_2022}, BMCTrack-d is most accurately described as ReID with tracking capabilities. For individual-level monitoring, it is of utmost importance to know the identities of all animals in the scene for as much of the time as possible. Given that the main goal of the algorithm is to assign a class ID per frame, it does not strictly enforce positionally consistent tracks. This means that if in frame $t$ class 1 is (wrongly) detected in the opposite corner of the pen from where it was detected in frame $t$-1, this does not lead to the initialisation of a new track. While this “identification first” approach leads to less strictly enforced positional consistency, it also facilitates self-correcting, which is the biggest strength of the proposed algorithm. BMCTrack-d might make mistakes more frequently because it heavily relies on a fallible classifier, but, because of the continuous nature of the classification, has the ability to self-correct these mistakes later. In simple words, traditional trackers might be right for a long time and then wrong for a long time, while BMCTrack-d might make mistakes earlier and more frequently, but correct the mistakes along the way. Figure \ref{fig:qualitative} illustrates this difference for pig dot horizontal line in test clip \textit{penA\_30s\_2}. BMCTrack-d (left image) repeatedly misdetected the pig in phases of crowding (top right corner) and fast movement (top right to bottom left diagonal track) but continuously corrected these mistakes. BoT-SORT (right image) was more robust to fast movement, however on losing the track in the crowded phase (top right corner) it was unable to recover the pig’s identity. While BoT-SORT’s ability to re-activate a lost track is limited to short intervals (30 frames by default), BMCTrack-d’s ability to re-identify an individual, in principle, is temporally unlimited. BMCTrack-d’s self-correction ability is absolutely crucial in the face of severe occlusions and motion blur, which frequently lead to lost tracks, and allows BMCTrack-d to outperform traditional trackers in the challenging scenario of this study.

The use of a separate classifier model comes at the expense of runtime efficiency. However, it is not uncommon in animal monitoring to operate at low frame rates (e.g., \citet{t_psota_long-term_2020}). At a frame rate of e.g., 6 fps, BMCTrack-d runs at real-time speed on the hardware used in this study. It is also quite robust to low frame rates, as shown in Table \ref{tab:6fps-comparison}, which compares the performance of BMCTrack-d to TrackTrack-ReID at 6 fps. BMCTrack-d’s performance remains more stable (-2.27\% HOTA) than TrackTrack-ReID’s (-9.35\% HOTA) compared to the original frame rate of 25 fps.

\begin{table}[h]
    \centering
    \caption{Comparison of the test set performance of BMCtrack-d and TrackTrack-ReID at 6 fps.}
    \label{tab:6fps-comparison}
    \resizebox{\columnwidth}{!}{%
        \begin{tabular}{lcccc}
            \toprule
            \textbf{method} & \textbf{LocA} & \textbf{AssA} & \textbf{DetA} & \textbf{HOTA} \\
            \midrule
            TrackTrack-ReID & \textbf{0.8869} & 0.6210          & 0.8260          & 0.6676          \\
            BMCTrack-d      & 0.8865          & \textbf{0.6687} & \textbf{0.8589} & \textbf{0.7487} \\
            \bottomrule
        \end{tabular}
    }
\end{table}

\subsection{BMCTrack-d modules}
\begin{figure*}[h]
  \centering
  \includegraphics[width=1.0\textwidth]{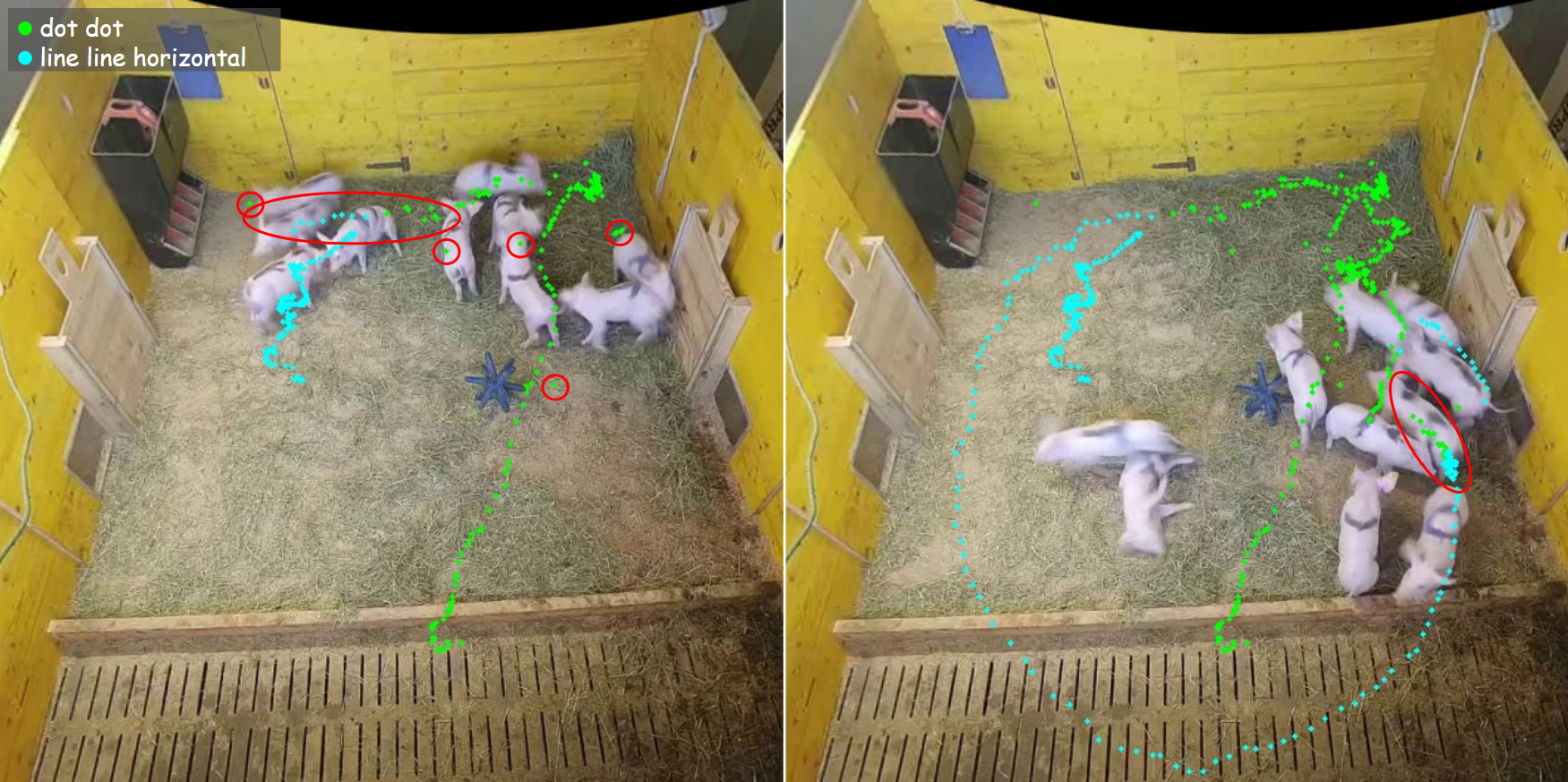}
  \caption{Illustration of the interplay between the temporal prediction consistency (TPC) check and the deduplication. The left image shows pig \textit{dot dot} was tracked with high accuracy for a long stretch and single misclassifications (small red circles) were corrected promptly by the TPC check. Then, on fast movement, dot dot was misclassified as \textit{line line horizontal} (red ellipse). The right image shows that this misclassification was corrected only after the pig slowed down (red ellipse). Presumably, the reduced motion blur and the resulting improved visibility of the back mark increased the classifiers confidence in the correct class, which was then reassigned by the deduplication step.}
  \label{fig:bmctrack}
\end{figure*}

Extensive experimentation showed both the TPC check as well as deduplication to play important roles in BMCTrack-d's performance. Deduplication could be shown to have clear benefits on its own (+12.93\% AssA, BMC vs. BMC-d), which is not the case for the TPC check (-3.18\% AssA, BMC vs. BMCTrack). Interestingly, the TPC check improved the performance only in combination with deduplication. This can be explained by viewing deduplication as a way of breaking out of a sequence of wrong predictions that the TPC check got stuck in. The TPC check works well for correcting singular wrong predictions but cannot correct sequences of wrong predictions (e.g., caused by a pig that remains partially occluded over an extended amount of time). These, however, are likely to cause class ID collisions at some point and will be corrected by deduplication, giving the TPC check a chance to reset. Figure \ref{fig:bmctrack} illustrates both ideas on test clip \textit{penA\_30s\_5}. In combination, TPC check and deduplication lead to a significant improvement in association (+26.74\% AssA, BMC vs. BMCTrack-d). As deduplication eliminates duplicate predictions, there was also an improvement in detection (+12.03\% DetA).

\subsection{Comparison to other trackers}
BoT-SORT updates ByteTrack in 3 ways: 1) enhanced Kalman filter, 2) camera motion correction and 3) ReID capabilities. As the (optional) ReID capabilities were investigated separately (BoT-SORT-ReID) and the camera in this study's setup was static, the improvement that BoT-SORT achieved over ByteTrack must be explained by the enhanced Kalman filter, which is plausible, because it alleviates the challenge that fast movement poses. Unfortunately, BMCTrack-d cannot benefit from Kalman filter forecasting, because the occasional jumps in location that BMCTrack-d allows are incompatible with the linear motion that Kalman filters assume. TrackTrack improves both these earlier tracking algorithms by assigning detections to existing trajectories from the track's perspective rather than solving a global assignment problem. This strategy is aimed at improving tracking performance in crowded scenes with a high number of occlusions and is especially relevant for pig tracking. The evaluation shows this algorithmic improvement to have a noticeable effect on the tracking performance, especially in combination with appearance-based reID (+11.23\% AssA, TrackTrack-ReID vs. BoT-SORT-ReID). Despite this innovation BMCTrack-d still outperforms TrackTrack on the tracking metrics (+1.03\% HOTA), while simultaneously solving the reID task by not just tracing but recognising the individual pigs.

\begin{figure*}[p]
  \centering
  \includegraphics[width=0.95\textwidth]{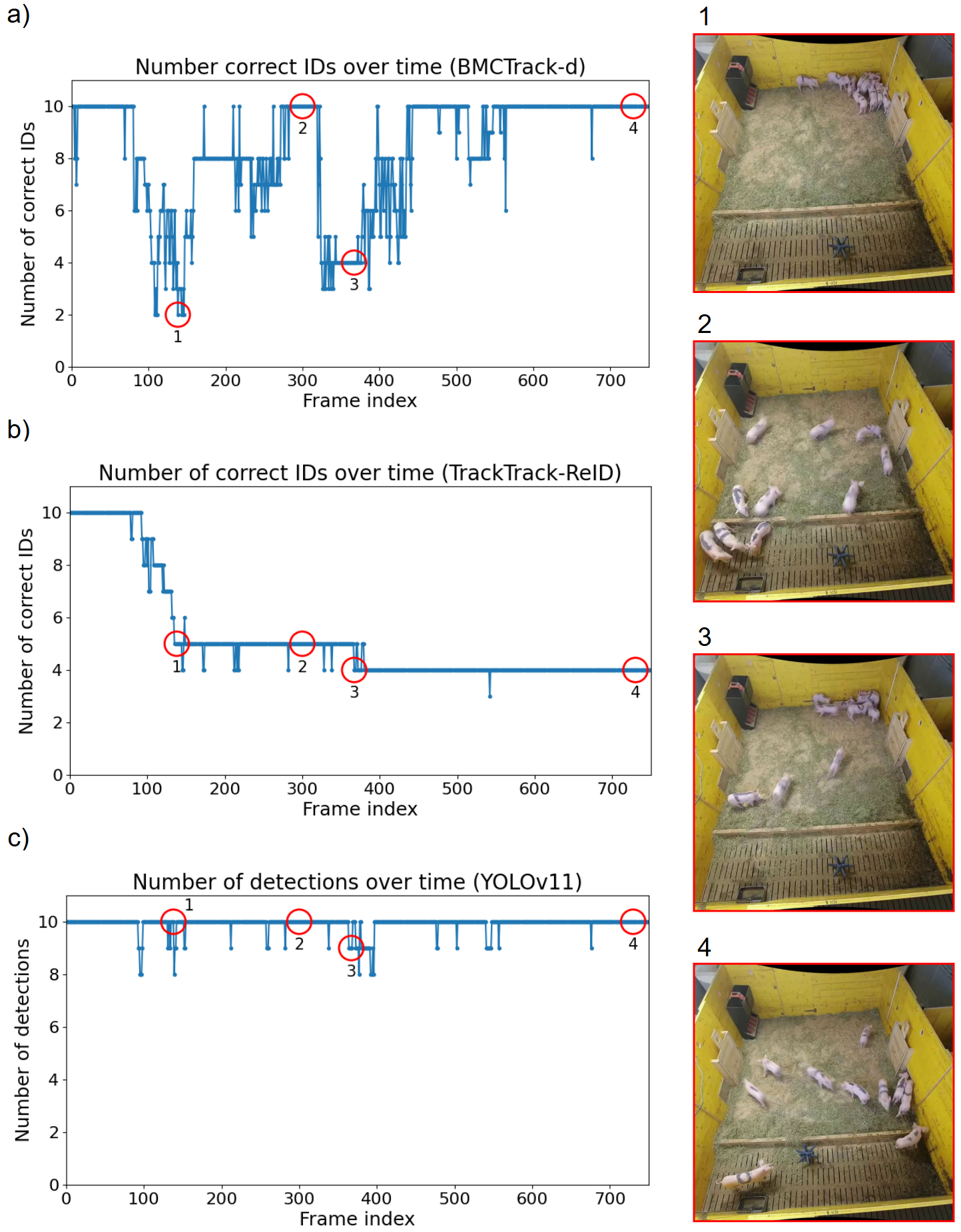}
  \caption{Per-frame comparison of BMCTrack-d (a) and TrackTrack-ReID (b) on test clip \textit{penA\_30s\_2}. Lost identities correlate with phases of high occlusion (c) for both methods. TrackTrack-ReID is capable of re-identifying lost individuals in cases of brief occlusions (indicated by the spikes), but unable to recover them in case of lasting occlusions. BMCTrack-d is able to fully recover all identities whenever the occlusion severity decreases.}
  \label{fig:strength}
\end{figure*}

The evaluation on the individual clips, shown in Tables \ref{tab:per-clip-bmctrack-d} and \ref{tab:per-clip-tracktrack-reid}, reveal a noticeable divergence in the performance of BMCTrack-d and TrackTrack-ReID on two of the test clips, \textit{penA\_30s\_2} (+27.82\% HOTA, BMCTrack-d vs. TrackTrack-ReID) and \textit{penA\_10s\_8} (-23.2\% HOTA, BMCTrack-d vs. TrackTrack-ReID). Deeper analysis shows that these cases align with the previously discussed strengths and weaknesses of BMCTrack-d. The clip \textit{penA\_30s\_2} shows a sequence of very rapid, erratic movement across the whole pen with high levels of motion blur, as well as low resolution pigs and severe occlusions whenever the pigs move to the far end of the pen. Fig. \ref{fig:strength}c shows that the performances of the trackers correlate with the occlusion severity (measured by the performance of the pig detector). To realise true individual-level monitoring, a tracker must be able to provide correct identity labels on the other side of such bouts of movement. Figure \ref{fig:strength}a shows that BMCTrack-d is capable of doing so. While TrackTrack-ReID is able to re-identify lost individuals after brief occlusions (Fig. \ref{fig:strength}b, indicated by the spikes in the curve), longer occlusions lead to permanently lost tracks. BMCTrack-d’s ability for reID is temporally unlimited, allowing it to fully obtain correct identity labels even after periods of severe occlusions.

\begin{figure*}[p]
  \centering
  \includegraphics[width=0.95\textwidth]{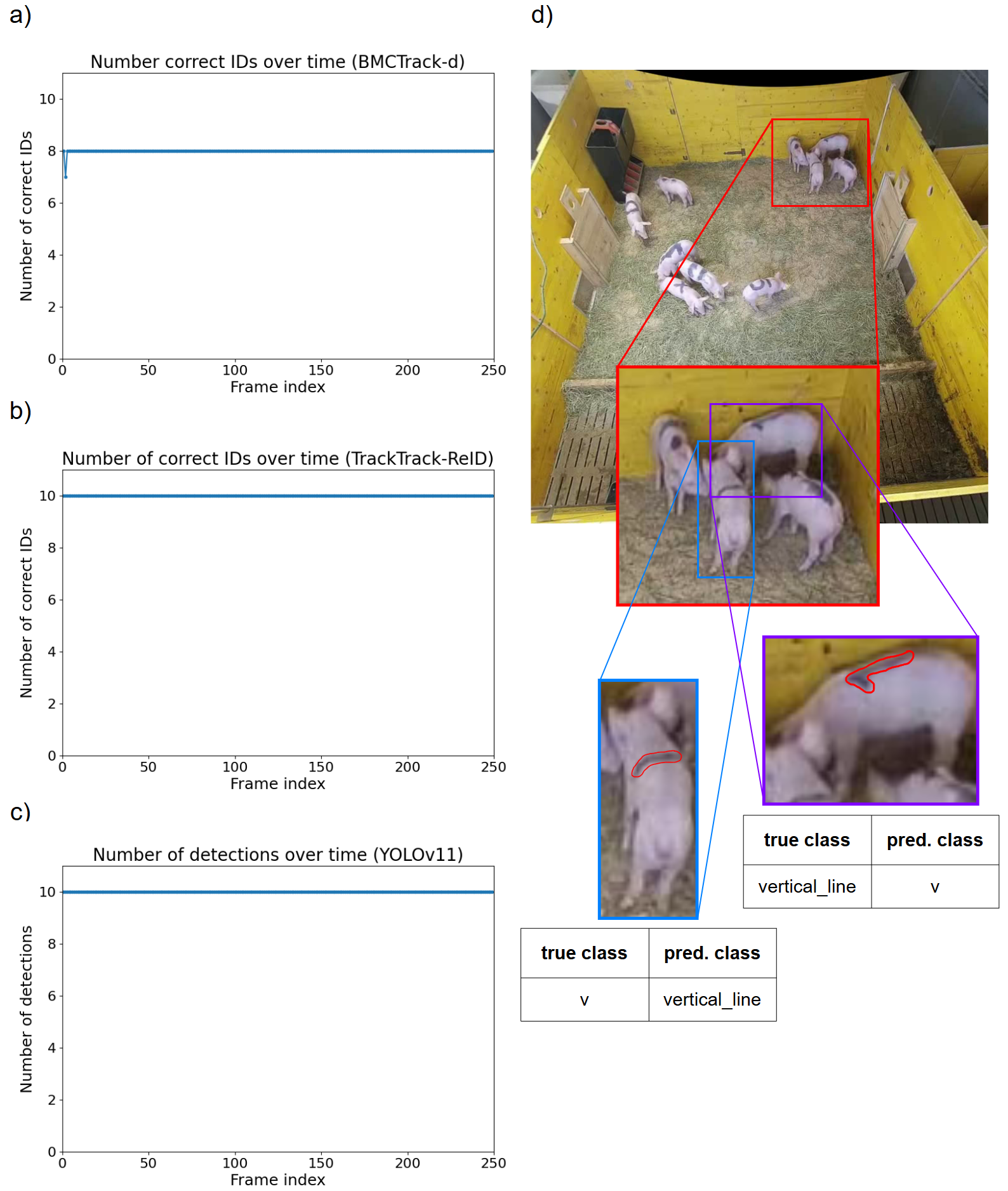}
  \caption{Per-frame comparison of BMCTrack-d (a) and TrackTrack-ReID (b) on test clip \textit{penA\_10s\_8}. TrackTrack-ReID perfectly tracks all individuals. BMCTrack-d confuses two individuals in absence of occlusions (c). Qualitative analysis shows that an imprecisely drawn back mark in combination with a specific view angle causes these back marks to resemble each other (d).}
  \label{fig:weakness}
\end{figure*}

\begin{figure*}[h]
  \centering
  \includegraphics[width=1.0\textwidth]{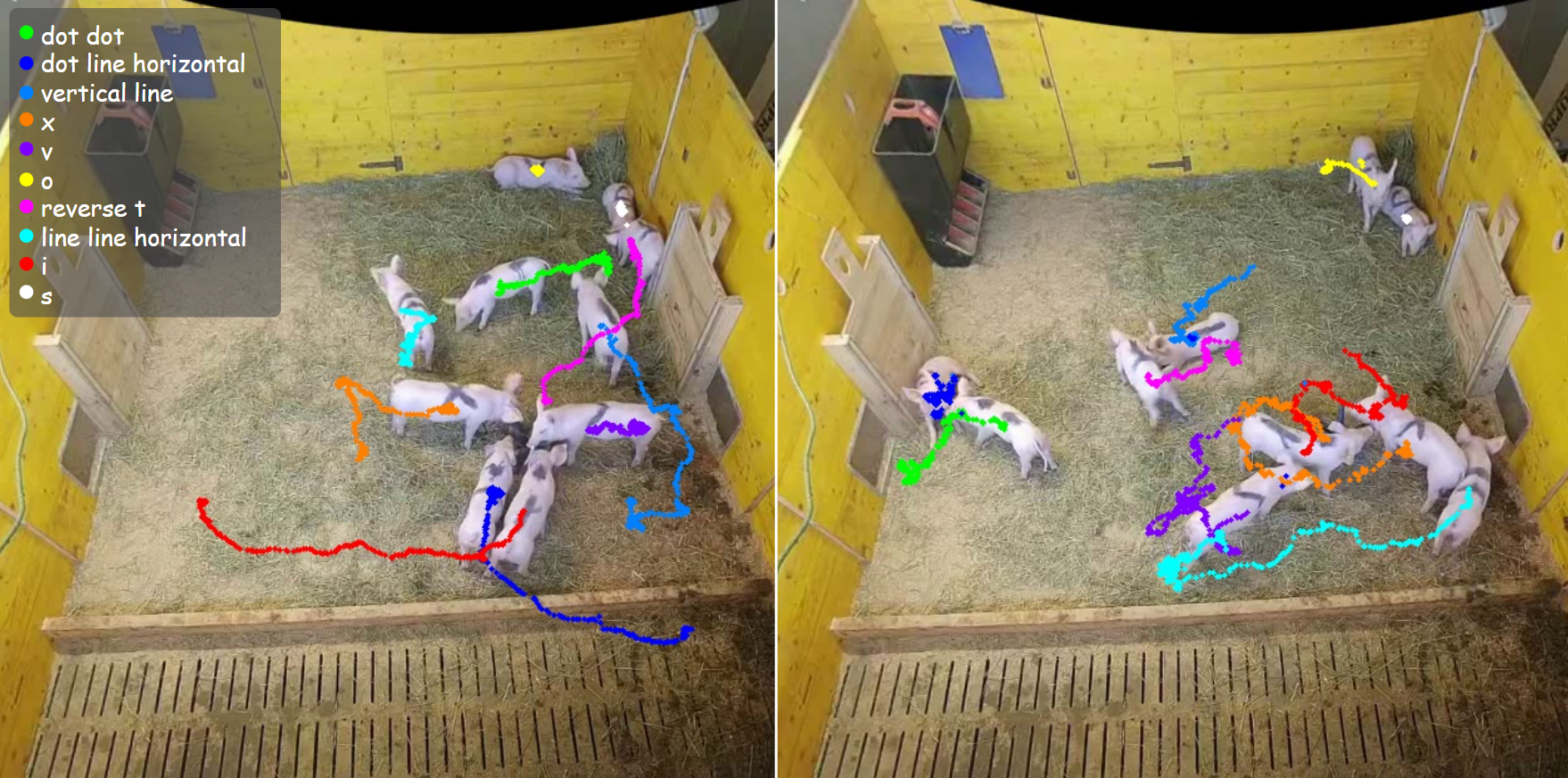}
  \caption{Qualitative results of BMCTrack-d on two test clips, \textit{penA\_10s\_6} (left) and \textit{penA\_10s\_7} (right).}
  \label{fig:visualization}
\end{figure*}

Conversely, in test clip \textit{penA\_10s\_8} BMCTrack-d consistently confuses two individuals (Fig. \ref{fig:weakness}a) Given that there are no notable occlusions (Fig. \ref{fig:weakness}c), the error must have a different cause. Qualitative analysis shows that the error most likely stems from an anomaly in the appearance of the back marks at the beginning of the clip, in which specific circumstances caused back marks \textit{vertical line} and \textit{v} to resemble each other. Imprecision in the drawing of the back mark \textit{vertical line} caused it to resemble back mark v. At the same time, v, at a specific view angle, resembled \textit{vertical line}, causing BMCTrack-d to flip the identity predictions, unable to correct this mistake for the rest of the video. This scenario shows that for the algorithmic use of back marks they must be carefully designed to minimise potential collisions \citep{brunner_insights_2026}. TrackTrack-ReID was unaffected by this and achieves perfect tracking throughout the whole clip. A quantitative analysis of the clips’ properties (resolution, occlusion, motion blur) can be found in Appendix \ref{app:vis}, Figures \ref{app:properties1} and \ref{app:properties2}. On all other test clips the results of both methods are comparable. Figure \ref{fig:visualization} shows qualitative results of BMCTrack-d on test clips \textit{penA\_10s\_6} and \textit{penA\_10s\_7}.

\subsection{Scope}
This study shows the potential of back marks for reID in challenging camera settings. Especially this robustness to non-idealised camera settings addresses an important gap in existing literature regarding practical application. However, while the non-idealised camera setting gives the results of this study high practical relevance, the experimental setup diverges from real-world farm settings in other important ways. At present, some of the experimental conditions in this study make BMCTrack-d more suitable for research settings and might limit its applicability to real-world farm settings. These limitations are discussed in detail in Section \ref{sec:limitations}.

\subsection{Limitations}
\label{sec:limitations}
\subsubsection{Runtime efficiency}
Farm settings might require running the tracking and all downstream tasks (such as behaviour recognition) directly on the camera stream, to allow real-time (online) intervention on detection of relevant behaviours. Research settings (such as observational studies) impose less strict requirements on algorithm efficiency, because the recording of the animals and the analysis of the data are two separate steps, of which the former might be finished before the latter is even started. The proposed tracking algorithm was predominantly developed for research settings. Therefore, runtime optimisation was not a priority and the algorithm in its current form might not be suited for online use. The back mark classification model architecture is a natural starting point for efficiency improvements, given that it is responsible for much of the processing time. Optimizing this architecture is left for future work.

\subsubsection{Closed-set settings}
BMCTrack-d was designed for closed-set applications, in which the set of possible identities is known beforehand, and these identities need to be continuously assigned to the correct individuals. This is a reasonable assumption for observational studies but might be prohibitive for many real-world farm settings. At present, BMCTrack-d is not able to dynamically handle the introduction of new back marks, that are not in the set of known identities. If a pen of size 10 is selected for individual-level monitoring and the classifier trained on 10 unique back marks, the post hoc addition of an 11th pig would require retraining the classifier. Conversely, the removal of pigs does not pose an issue. From the perspective of a tracker, situations in which not all known pigs are present arise regularly, namely whenever occlusions occur. These scenarios are covered by the test set, which includes clips with severe occlusions (e.g., \textit{penA\_30s\_2}) and the evaluation shows that BMCTrack-d can handle such scenarios. The permanent removal of a pig is out of scope for the current study but could be handled based on its effects, which would be twofold: 1) the mean number of pig detections would drop (e.g., from 10 to 9) and 2) the affected back mark would appear less frequently in the predictions (only in case of mistakes). Thus, this situation could be handled automatically, by 1) detecting the removal via a drop in the mean number of detections, and 2) removing the affected back mark from the list of known back marks (i.e., ignoring its prediction). If the pig is reintroduced to the pen at a later point in time, the process can be reversed, i.e., detecting an increase in the mean number of detections and reactivating the back mark in the list. A practical evaluation of this is left for future work.

\subsubsection{Back mark application}
This study showed that back marks are a viable means for pig reID in challenging settings. However, the use of back marks comes with important practical challenges. In this study, the back marks were applied manually and refreshed 1-2 times a week. While this is a feasible option for research, it might be prohibitive for many real-world farm settings, which might require automatic application of the back marks.  In some practical farm settings, e.g., pig units for genetic selection and performance evaluation, manual application of back marks might be feasible as well. Genetic evaluation is a highly resource-intensive operation compared to standard pork production, where pigs are weighed individually or ultrasound scans are performed routinely on live animals to measure backfat thickness and muscle depth. Camera based individual phenotyping might allow breeding organizations to select for complex traits recorded continuously, objectively and in high-frequency. These traits e.g., aggression in pigs, were previously impossible to be measured accurately. Higher expenses on such units, which might be related to individual back marking, might be justified as these units function as centralized hubs for accelerating genetic progress across the industry.  Manual application of individual back marks on standard commercial farms might be less feasible. However, there are commercial products, which offer automated solution e.g., Colortek (Fancom, Panningen, the Netherlands) for sow spray marking inside the feeding stations, which supports the farmers in identifying sows in heat or to provide information on their health status. Further development and adaptations of these systems might support practicality of using back marks for automated identification of pigs with BMCTrack-d. Development of BMCTrack-d in a pen with 10 pigs supports its practical use in conventional European finishing systems which commonly use small group sizes with 10 pigs e.g., in the Netherlands or Sweden. However, validation in larger group sizes is needed.

\section{Conclusion}
\label{sec:conclusion}
Reliable reID and tracking are imperative for individual-level pig monitoring. This study proposed a novel tracking algorithm, which addresses two important challenges towards this goal, the difficulty of differentiating individuals of uniform appearance and the difficulty of tracking in challenging situations caused by experimental conditions and pig behaviour. BMCTrack-d demonstrates the merit of back marks for differentiating the pigs and shows high robustness when confronted with practical challenges like occlusions, motion blur and low-resolution recordings. Especially its ability to recover lost identities, which is crucial for downstream tasks like behaviour recognition, sets it apart from existing solutions. On a challenging side-view dataset, BMCTrack-d outperformed BoT-SORT-ReID and TrackTrack-ReID by 9.11\% and 1.03\% HOTA, respectively, while simultaneously solving the reID task by not just tracing but recognising the individual pigs. Subsequent studies should broaden the spectrum of possible applications of BMCTrack-d, by improving runtime efficiency via more specialised back mark classifiers, reviewing and optimising the back mark design for easier identification, and evaluating automated back mark application technologies.

\section*{Acknowledgments}
This research was funded in whole or in part by the Austrian Science Fund (FWF) [https://doi.org/10.55776/DFH34]. For open access purposes, the author has applied a CC BY public copyright license to any author-accepted manuscript version arising from this submission.
The data used in this study originates from the "Let me out" project, funded by the Austrian Science Fund (FWF), project I 6488-B [https://doi.org/10.55776/I6488].
Further, we would like to thank Janina Weißenborn and Stefan Kupfer from the University of Veterinary Medicine Vienna for help with the data annotation and technical support, respectively.

\section*{Author Contributions}
\textbf{David Brunner}: Conceptualization, Data curation, Investigation, Methodology, Software, Validation, Visualization, Writing – original draft. \textbf{Maciej Oczak}: Conceptualization, Funding Acquisition, Methodology, Resources, Writing – review \& editing. \textbf{Marie Bordes}: Data curation, Resources, Writing – review \& editing. \textbf{Jean-Loup Rault}: Funding acquisition, Resources, Writing – review \& editing. \textbf{Stephan M. Winkler}: Funding acquisition, Supervision, Writing – review \& editing. \textbf{Viktoria Dorfer}: Funding acquisition, Project administration, Supervision, Writing – review \& editing.

\section*{Ethics statement}
All methods and animal use were approved by the Animal Ethics Committee of the University of Veterinary Medicine, Vienna (reference number 2024-0.026.412), and carried out in accordance with Good Scientific Practice guidelines and national legislation.


\appendix
\renewcommand{\thetable}{\Alph{section}.\arabic{subsection}.\arabic{table}}
\counterwithin{subsection}{section}
\counterwithin{table}{subsection}
\counterwithin{figure}{subsection}

\section{Tracking metrics}
\label{app:metrics}
The metric adopted for the evaluation of the tracking algorithms in this work is the higher order tracking accuracy (HOTA) \citep{luiten_hota_2021}. It improves upon the most important traditional tracking metrics, multi-object tracking accuracy (MOTA) \citep{bernardin_evaluating_2008} and identification F1 (IDF1) \citep{ristani_performance_2016}, in many meaningful ways, which are criticised for putting too much emphasis on detection and association, respectively.
To calculate HOTA, first, the detections of the tracker need to be matched to the ground truth bounding boxes. An important distinction in tracking evaluation is that between matches and associations. A match describes a situation in which the IoU between a detected bounding box and a ground truth bounding box in a frame exceeds a defined threshold. An association describes when the detected and the ground truth bounding box of a match also have the same (class) ID. The best pairing of detections and ground truths is searched, so that across the whole video the best final HOTA score is achieved. This only implicitly enforces connected trajectories – the probability that the best overall score is attained by matching the same ground truth to detections with different (class) IDs in subsequent frames is low, because for every match (in a given frame) the consequences for the whole video are checked. In addition to this association score, every match also has a localisation similarity, defined by the IoU between the detections. The Hungarian algorithm is used to maximise 1) the total number of matches, 2) the mean association score and 3) the mean localisation similarity. Once the best pairing is found, $HOTA_\alpha$ is calculated as:
\begin{equation}
    \begin{aligned}
        HOTA_{\alpha} &= \sqrt{\frac{\sum_{c \in \{TP\}} A(c)}{|TP|+|FP|+|FN|}} \\
        A(c) &= \frac{|TPA(c)|}{|TPA(c)|+|FPA(c)|+|FNA(c)|}
    \end{aligned}
\end{equation}
where $c$ is a given match and $|TP|$ is the total number of matches in the optimised pairing. $|FP|$ and $|FN|$ are the total number of resulting false positives and false negatives, respectively. $|TPA|$ is the number of true associations, i.e. the number of correct ID matches between ground truths and detections of the same ID (along the trajectory) that result from this match. $|FPA|$ and $|FNA|$ are the number of false positive associations and false negative associations, respectively. 
A great advantage of HOTA is that it can be decomposed into sub-metrics that allow evaluating a tracker’s performance on each of the capabilities involved in successful tracking separately, namely detection accuracy (DetA) and association accuracy (AssA):
\begin{equation}
    DetA_{\alpha} = \frac{|TP|}{|TP|+|FN|+|FP|}
\end{equation}
\begin{equation}
    AssA_{\alpha} = \frac{1}{|TP|}\sum_{c \in \{TP\}} A(c)
\end{equation}
where
\begin{equation}
    HOTA_{\alpha} = \sqrt{DetA_{\alpha} \cdot AssA_{\alpha}}
\end{equation}
The localisation accuracy (LocA) can be calculated separately, as:
\begin{equation}
    LocA_{\alpha} = \frac{1}{|TP_{\alpha}|}\sum_{c \in \{TP_{\alpha}\}} S(c)
\end{equation}		                  (5)
where $S$ is the localisation similarity, measured as the IoU between detection and ground truth.
$HOTA_\alpha$ is calculated for every localisation similarity threshold $\alpha \epsilon L$, and the average gives the final score: 
\begin{align}
    HOTA &= \frac{1}{|L|}\sum_{\alpha \in L} HOTA_{\alpha} \\
    L &= \{0.05, 0.1, \ldots, 0.9, 0.95\} \notag
\end{align}

\section{Additional results}
\label{app:additional}
\subsection{Extended optimal configuration study}
This section discusses additional experiments on the optimal configuration of BMCTrack-d. First, the optimal frame buffer length for the TPC check is assessed in the range $[2, 6]$. Next, alternative matching variants are evaluated, namely BMCTrack-d-OKS, which uses keypoint-skeleton-based matching, as well as BMCTrack-d-KP, which supplements IoU-based matching with orientation information derived from two keypoints. The results of the study on the optimal frame buffer length are summarised in Table \ref{tab:frame-buffer-length} and show number of last frames $n=4$ to be the best setting. Table \ref{tab:enhanced-matching-variants} shows the results of the enhanced matching variants study.

\begin{table}[h]
    \centering
    \caption{The results of the frame buffer length study. Best results in bold.}
    \label{tab:frame-buffer-length}
    \resizebox{\columnwidth}{!}{%
        \begin{tabular}{lcccc}
            \toprule
            \textbf{frame buffer length} & \textbf{LocA} & \textbf{AssA} & \textbf{DetA} & \textbf{HOTA} \\
            \midrule
            2 & 0.8864 & 0.6737          & 0.8591 & 0.7529          \\
            3 & 0.8864 & 0.6821          & 0.8591 & 0.7570          \\
            4 & 0.8864 & \textbf{0.7048} & 0.8591 & \textbf{0.7714} \\
            5 & 0.8864 & 0.6608          & 0.8591 & 0.7459          \\
            6 & 0.8864 & 0.6590          & 0.8591 & 0.7448          \\
            \bottomrule
        \end{tabular}
    }
\end{table}

The frame buffer length represents a trade-off between detection matching and prediction correction. The longer the frame buffer, the more likely the matching fails. Matching detections between frame $t$ and frame $t$-2 is more likely to produce correct matches than between frame $t$ and frame $t$-5, because the potential offset through movement is smaller. At the same time, the majority vote strategy of the TPC check benefits from longer sequences of class predictions, because, assuming that wrong predictions are in the minority, this minority becomes clearer in longer sequences. A frame buffer length of $n=4$ seems to optimise this trade-off, as shown in Table \ref{tab:frame-buffer-length}.

\begin{table}[h]
    \centering
    \caption{The results of the enhanced matching variants study.}
    \label{tab:enhanced-matching-variants}
    \resizebox{\columnwidth}{!}{%
        \begin{tabular}{lcccc}
            \toprule
            \textbf{method} & \textbf{LocA} & \textbf{AssA} & \textbf{DetA} & \textbf{HOTA} \\
            \midrule
            BMCTrack-d       & 0.8864 & \textbf{0.7048} & 0.8591 & \textbf{0.7714} \\
            BMCTrack-d-OKS   & 0.8864 & 0.6608          & 0.8591 & 0.7461          \\
            BMCTrack-d-KP    & 0.8864 & 0.6879          & 0.8591 & 0.7628          \\
            \bottomrule
        \end{tabular}
    }
\end{table}

\begin{figure*}[h]
  \centering
  \includegraphics[width=0.70\textwidth]{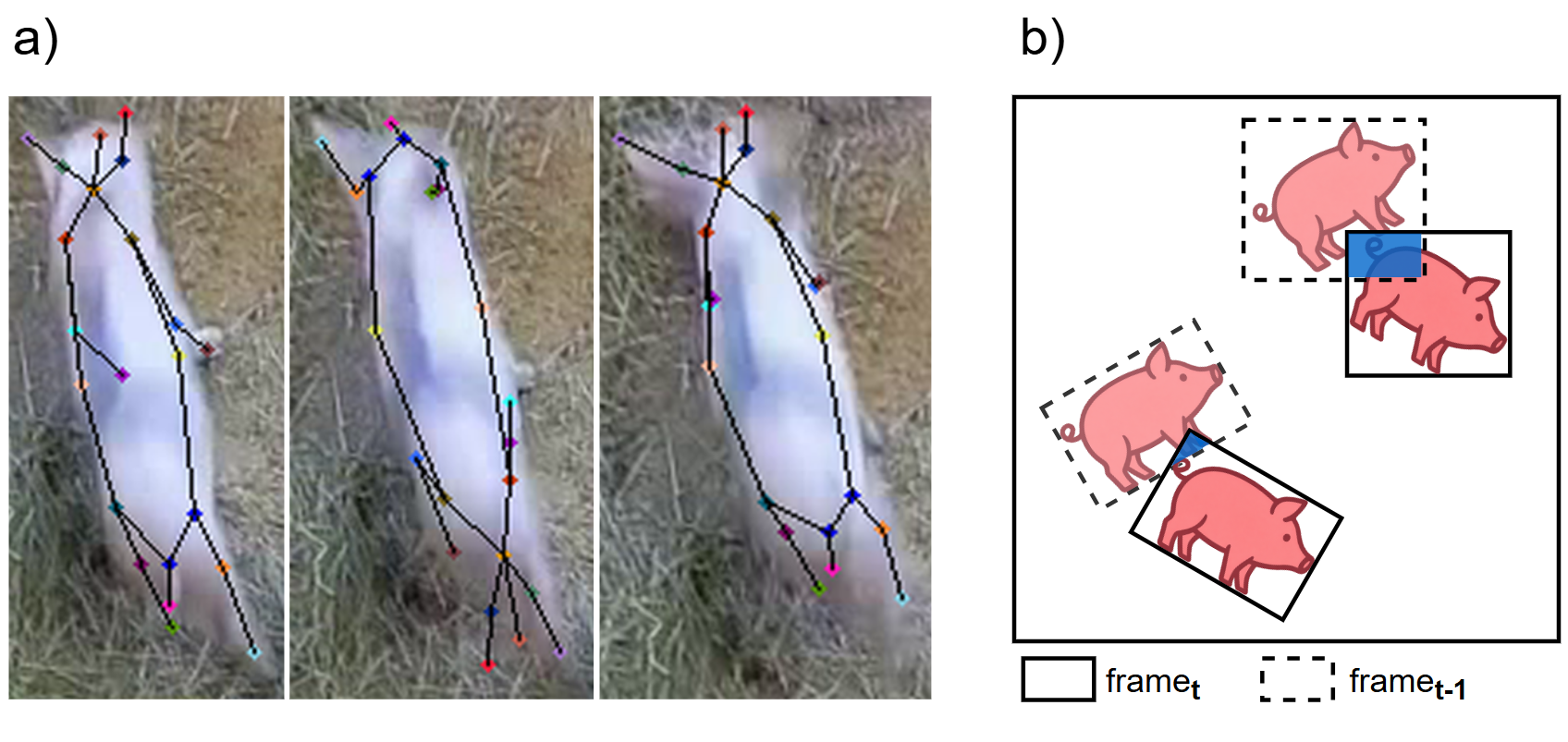}
  \caption{Illustration of the skeleton flip issue of the pose estimation model (a) and the reduced overlap of oriented bounding boxes in consecutive frames (b).}
  \label{app:flipobb}
\end{figure*}

Neither substituting IoU-based matching with OKS-based matching (BMCTrack-d-OKS), nor supplementing the IoU-based matching with keypoint-based orientation information (BMCTrack-d-KP) could improve performance. A likely explanation for this is that both require near-perfect pose estimation to work as intended.  While the pose estimation model used in this work reaches high accuracy, it suffers from occasional pose inversions, in which the keypoint skeleton flips along a pig’s body for single frames, as depicted in Fig. \ref{app:flipobb}a. These flips completely disrupt the matching process and demonstrate the fragility of keypoint-based tracking. To study the theoretical benefits of these methods, ground truth keypoint skeletons are required, which were not available for the data in this study. This evaluation is left for future work.

\subsection{Oriented bounding boxes}
\citet{odo_re-identification_2025} have reported improvements in tracking and reID on using oriented instead of axis-aligned bounding boxes. To establish the benefit of oriented bounding boxes for BMCTrack-d an instance of the YOLOv11 object detector for oriented bounding box detection is trained. The oriented bounding box labels are created by passing the training, validation and test datasets (described in Section \ref{sec:datasets}) to an instance of the Segment Anything Model \citep{kirillov_segment_2023} and fitting rectangles on the generated segmentations, which represents a slightly simplified version of the strategy proposed in \citet{odo_automated_2024}. Given that the generated oriented bounding boxes can contain imprecisions and errors, the test set was manually reviewed and frames with deficient bounding boxes were filtered out. YOLOv11-OBB, trained as described above, reached 99.27\% mAP@0.5 (YOLOv11: 99.42\%) on the validation set. For a fair comparison of the resulting BMCTrack-d-OBB to the baseline BMCTrack-d, the test set of the latter was reduced to the same set of frames. The more precisely fitting oriented bounding boxes also harbour benefits for classification, because they show less background and are less likely to include multiple pigs. For this reason, a separate instance of the image classifier was trained on oriented bounding boxes for this experiment, reaching 94.86\% accuracy (axis-aligned classifier: 91\%) on the validation set. Table \ref{tab:oriented-bounding-box} shows the results.

\begin{table}[h]
    \centering
    \caption{The results of the oriented bounding box study. Best results in bold.}
    \label{tab:oriented-bounding-box}
    \resizebox{\columnwidth}{!}{%
        \begin{tabular}{lcccc}
            \toprule
            \textbf{method} & \textbf{LocA} & \textbf{AssA} & \textbf{DetA} & \textbf{HOTA} \\
            \midrule
            BMCTrack-d     & \textbf{0.8889} & \textbf{0.7366} & \textbf{0.8744} & \textbf{0.7963} \\
            BMCTrack-d-OBB & 0.8583          & 0.6546          & 0.8151          & 0.7250          \\
            \bottomrule
        \end{tabular}
    }
\end{table}

Contrary to existing work (e.g., \citet{odo_re-identification_2025}), in this study, the use of oriented bounding boxes did not lead to improved tracking performance. Table \ref{tab:oriented-bounding-box} shows the drop in performance to be most noticeable for association (AssA -8.20\%). A possible explanation is that axis aligned bounding boxes promote the backwards matching in the TPC check, because the, on average, more expansive bounding box areas lead to greater overlaps and make matches more robust to movement, as illustrated in Fig. \ref{app:flipobb}b. For the more precise oriented bounding boxes, matching across frames frequently fails, especially if the pigs move rapidly.

\bibliography{references}


\renewcommand{\thefigure}{%
  \ifnum\value{subsection}=0
    \Alph{section}.\arabic{figure}%
  \else
    \Alph{section}.\arabic{subsection}.\arabic{figure}%
  \fi
}

\clearpage

\section{Additional visualisations}
\label{app:vis}
\mbox{}\\\\\\\\\\\\

\begin{figure}[h]
  \includegraphics[width=\columnwidth]{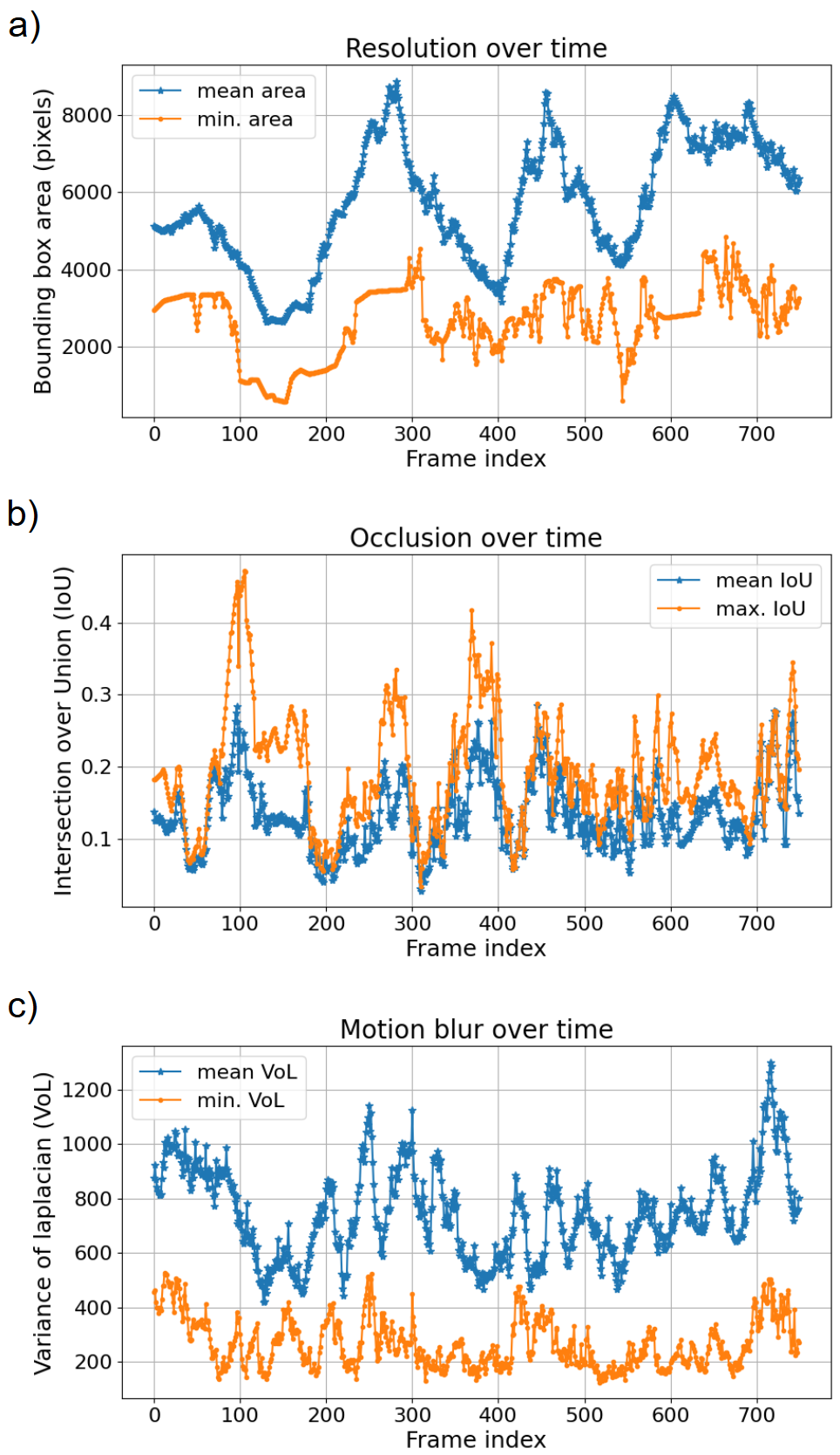}
  \caption{Scene property study for test clip \textit{penA\_30s\_2}. The resolution of the pigs’ bounding boxes (a) reflects the pigs’ movement towards and away from the camera. The distance to the camera also correlates with the level of occlusion as measured by the bounding box overlap (b), because the sharper angle at distance makes occlusions more likely. The variance of Laplacian (VoL) (c) is lowest both when the motion blur is highest, as well as when the resolution is lowest. These confounding effects are introduced by the side-view setting.}
  \label{app:properties1}
\end{figure}

\begin{figure}[h]
  \includegraphics[width=\columnwidth]{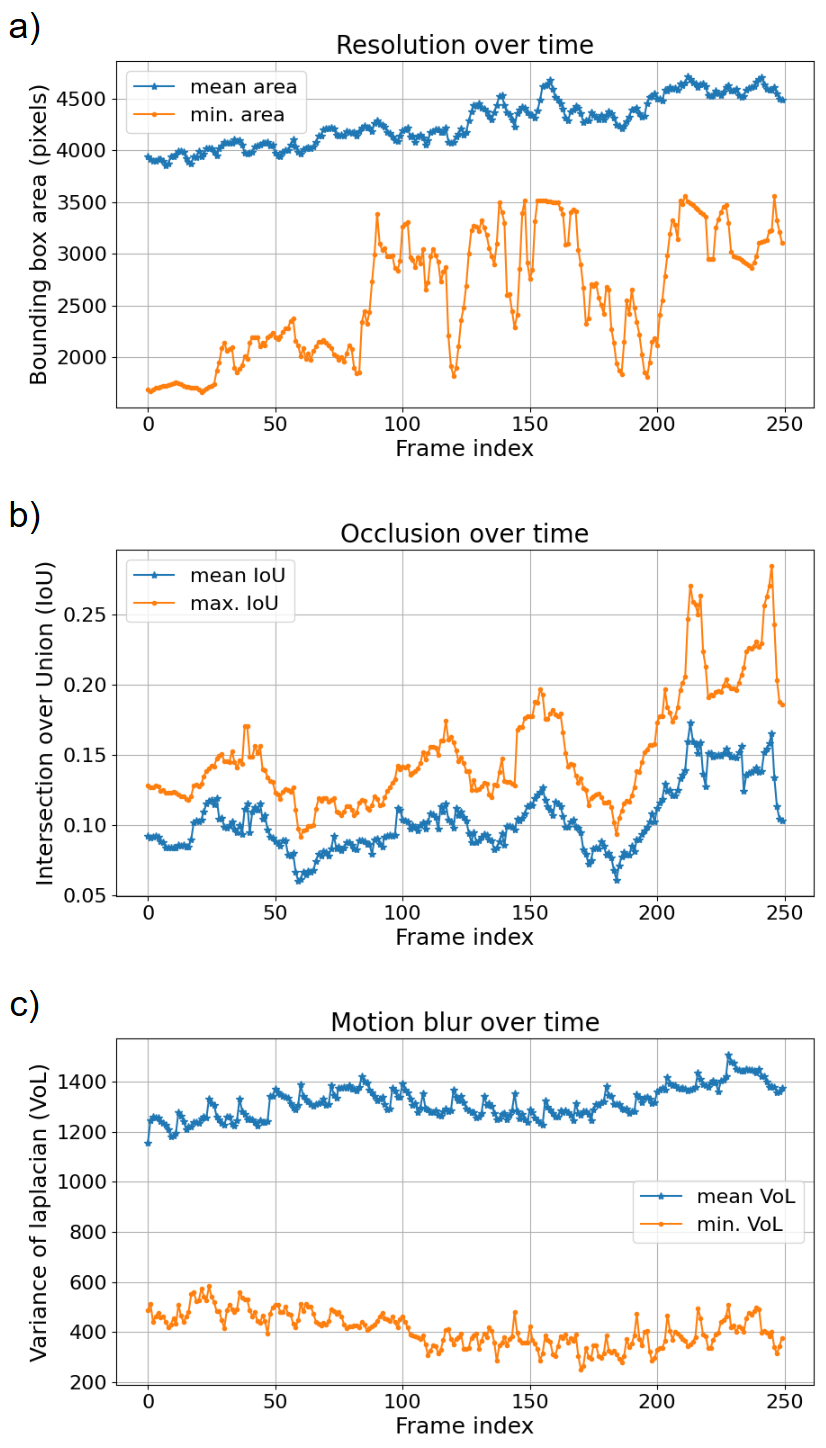}
  \caption{Scene property study for test clip \textit{penA\_10s\_8}. The increasing mean resolution points at a movement tendency towards the camera (a). The divergence in the mean and max./min. values of the properties reflects the fact that in this clip only individual pigs move while the others are stationary. The moving pigs cause occlusions towards the end of the clip (b). The constant VoL reflects the slow movement of the pigs (c).}
  \label{app:properties2}
\end{figure}


\begin{figure*}[p]
  \centering
  \includegraphics[width=0.9\textwidth]{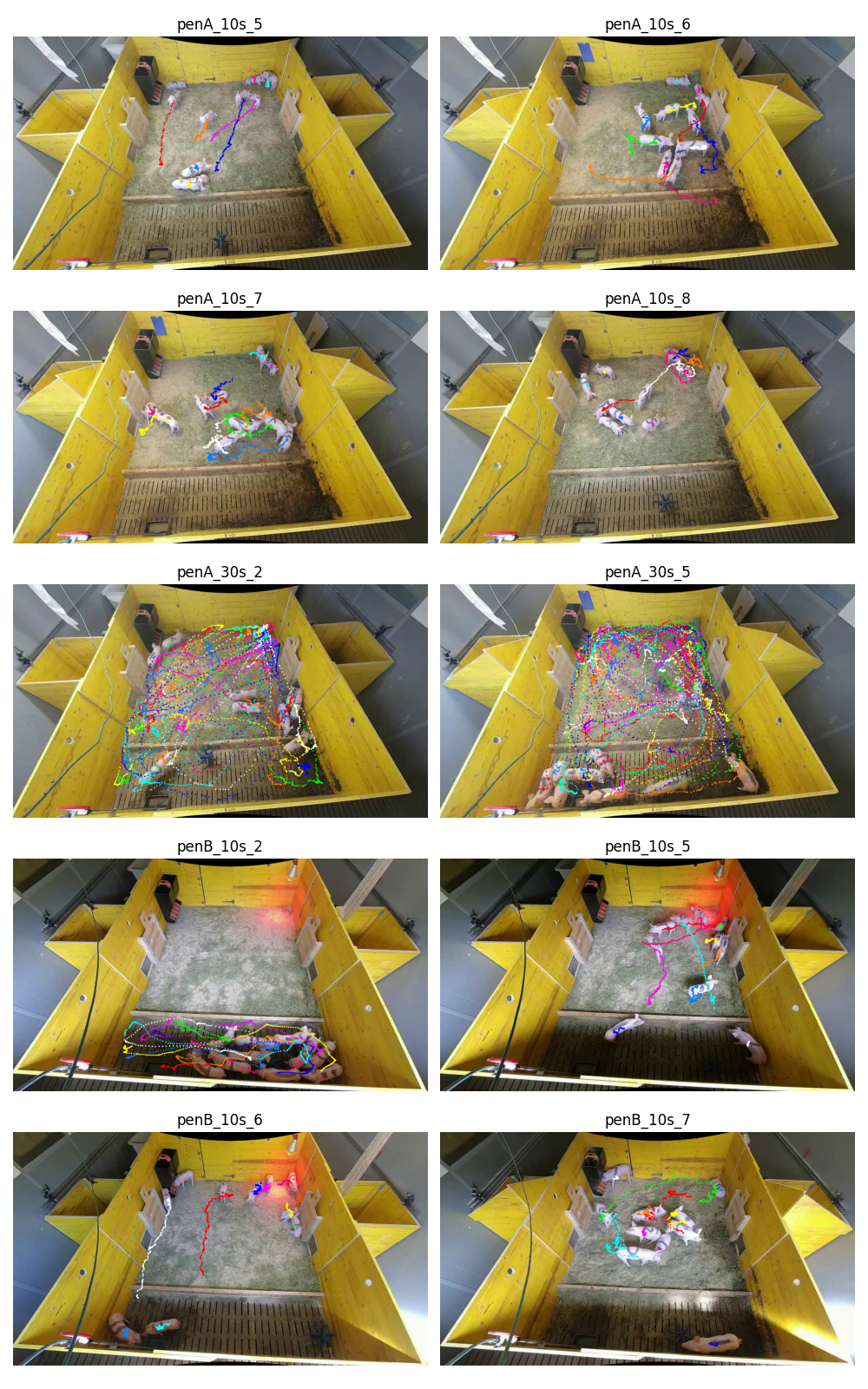}
  \caption{Visualisation of the pigs’ movement patterns in the test data clips.}
  \label{app:datatracks}
\end{figure*}

\end{document}